\documentclass[HARVARD,LATO2COL]{WileyNJDv5}

\articletype{Article Type}%

\copyyear{2026}
\startpage{1}

\usepackage{subcaption} 
\usepackage{todonotes}
\usepackage{multirow}
\usetikzlibrary{shapes.geometric, arrows, fit}

\begin{document}

\title{Reducing Barriers to Academic Support: Evaluating a Course-Specific RAG System for Addressing Help-Seeking Disparities in Higher Education}  

\author[1]{Andy Gray}

\author[1]{Jake Hobbs}

\titlemark{Reducing Barriers to Academic Support: Evaluating a Course-Specific RAG System for Addressing Help-Seeking Disparities in Higher Education}

\address[1]{
    \orgdiv{School of Design}, 
    \orgname{Bath Spa University}, 
    \orgaddress{\state{Bath}, 
    \country{UK}}
}


\abstract[Abstract]{
    Access to academic support is a key determinant of student success, yet students do not experience that access equally. While some students readily seek assistance from lecturers or tutors, others hesitate due to anxiety, fear of judgement, uncertainty about expectations, or low confidence in their own understanding. These help-seeking disparities may be particularly evident in computing education, where programming tasks are cumulative and cognitively demanding. Although students increasingly turn to general-purpose generative AI tools for assistance, such systems can produce responses that are inaccurate, insufficiently contextualised, or misaligned with module expectations.
    This study presents and evaluates Beacon, a course-specific Retrieval-Augmented Generation (RAG) system designed to provide private, immediate, and module-aligned academic support. Grounding responses exclusively in approved teaching materials, Beacon was designed to complement existing educational support by lowering barriers to help-seeking while encouraging independent learning. Using a Design-based research approach the study followed an iterative design process to develop Beacon with analysis conducted through a mixed-methods evaluation combining questionnaires with semi-structured interviews with both students and staff at a Higher Education institution. Students consistently described Beacon's responses as closely aligned with module content and more trustworthy than unrestricted generative AI tools, valuing its use of pseudocode and scaffolded explanations over direct solutions. Although participants remained appropriately cautious about trusting AI-generated responses without verification, they consistently viewed the system as a valuable first point of support before consulting lecturers or official learning resources.
    The findings suggest that carefully designed course-specific AI systems may reduce barriers to academic support by occupying an intermediary space between independent study and formal academic support. Rather than replacing educators, educational AI may be most valuable when it broadens access to academically appropriate guidance while preserving the pedagogical role of lecturers.
}


\jnlcitation{\cname{%
\author{Gray A.},
\author{Hobbs J.},
\ctitle{Reducing Barriers to Academic Support: Evaluating a Course‐Specific RAG System for Addressing Help‐Seeking Disparities in Higher Education.} \cjournal{\it Preprint. Not yet peer reviewed.} \cvol{2026;00(00):1--29}.}
}

\maketitle

\section*{Context and Implications}
    \subsection*{Rationale for this study}

        Students do not experience access to academic support equally. While universities provide a range of formal support mechanisms, many students remain reluctant to seek assistance due to anxiety, fear of judgement, low confidence, or uncertainty about where to ask for help. As a result, students increasingly turn to general-purpose large language models (LLMs), such as ChatGPT, as an accessible source of immediate academic support. Although these systems are readily available, they frequently generate responses that are detached from module-specific learning outcomes, institutional expectations, and assessment requirements, potentially creating confusion, reinforcing misconceptions, or encouraging inappropriate use.

        This study therefore investigates whether a course-specific Retrieval-Augmented Generation (RAG) system can provide a more educationally appropriate form of AI-supported learning. By grounding responses in approved teaching materials, the system aims to provide timely, trustworthy, and context-aware academic support that complements existing teaching practices while reducing barriers to help-seeking. In doing so, the approach also has the potential to promote academic integrity by encouraging students to engage with module-aligned guidance rather than relying on unrestricted external AI systems.
        
    \subsection*{Why the new findings matter}

        The findings demonstrate that barriers to academic support extend beyond the simple availability of support services. Students' willingness to seek help is shaped by factors such as confidence, anxiety, fear of judgement, and perceptions of whether their questions are appropriate to ask in staff-mediated or public learning environments. By providing private, immediate, and course-specific guidance, Beacon offered students a low-friction means of accessing support before engaging with lecturers or tutors. Rather than replacing existing support structures, the system occupied an intermediary space between independent study and formal academic support, lowering the threshold for help-seeking while preserving the central role of educators.
        
        The study also contributes to wider debates surrounding the responsible and equitable integration of generative AI in higher education. Although unrestricted AI systems can increase access to academic assistance, they may also reinforce educational inequalities when students receive responses that are inaccurate, insufficiently contextualised, or beyond the intended level of study. The findings suggest that course-specific Retrieval-Augmented Generation (RAG) offers one possible approach to addressing these challenges by grounding responses in approved teaching materials, promoting transparency, and encouraging scaffolded learning rather than answer substitution. More broadly, this work demonstrates how educational AI can be designed not simply to improve efficiency, but to widen equitable access to academically appropriate learning support.

     \subsection*{Implications}

        The findings have implications for educators, institutions, and researchers seeking to integrate generative AI into higher education in ways that promote equitable access to learning support. For educators, the study suggests that course-specific AI systems can complement existing teaching practices by providing students with immediate, module-aligned guidance outside scheduled teaching sessions. Rather than replacing lecturers or tutors, such systems may act as an accessible first point of support, encouraging students to build confidence before engaging with formal academic assistance.
        
        For higher education institutions, the findings demonstrate the importance of designing AI systems around pedagogical objectives rather than technological capability alone. Grounding responses in approved teaching materials, maintaining transparency regarding source information, and aligning AI outputs with module learning outcomes offer one approach to supporting responsible AI adoption while preserving trust in institutional teaching practices. Although course-specific AI is not a substitute for effective teaching or student support services, it may help reduce barriers to help-seeking for students who are less likely to access conventional forms of academic support.
        
        More broadly, this study contributes to the growing body of research on educational AI by illustrating how generative AI can be designed to address educational challenges rather than simply automate existing practices. The findings encourage future research to investigate how similar approaches perform across different disciplines, student populations, and institutional contexts, and whether such systems can contribute to longer-term improvements in student engagement, confidence, and educational equity.

\section{Introduction}
    \label{sec:intro}

    Access to academic support is not experienced equally across higher education (HE). Concerns about judgement, uncertainty about where to ask for help, and limited access to academic staff can discourage students from seeking assistance. These challenges are particularly pronounced in large classes where opportunities for individual support may be limited.

    In this paper, help-seeking disparity refers to unequal patterns of access to academic support that arise not only from the availability of support services, but also from students' confidence, prior experience, anxiety, fear of judgement, and ability to navigate institutional systems. Within computing education, these disparities may be particularly pronounced because programming difficulties are often highly individual, cumulative, and difficult for students to articulate. As a result, students who are less confident or more anxious may delay seeking help, rely on unsuitable external resources, or disengage from learning activities. These confidence and anxiety-related barriers to help-seeking are not evenly distributed across the student population. Prior research associates elevated anxiety and reduced confidence in help-seeking with mature-student \citep{chapman2017using}, first-generation or widening-participation status \citep{koh2022self} and gender and cultural background \citep{bornschlegl2020variables, ruihua2025understanding}. Help-seeking disparity, as examined in this study, may therefore intersect with, rather than sit apart from the demographic and identity-based disparities that generative AI (GenAI) in education is increasingly expected to address \citep{james2024levelling, ni2026mapping}.

    In recent years, students have increasingly turned to GenAI tools such as \textit{ChatGPT} to assist with their learning. While these systems can provide rapid responses to questions, they often generate explanations that are misaligned with course-specific materials or that assume levels of prior knowledge beyond those of novice learners. This can lead to confusion or reinforce misunderstandings. Furthermore, they may introduce new disparities by advantaging students who are already able to evaluate AI-generated responses critically, while disadvantaging those who are less confident or less able to identify inaccuracies.
    
    Retrieval-Augmented Generation (RAG) offers a potential solution to this challenge by grounding large language model (LLM) responses in curated sources of information. By retrieving relevant content from course materials before generating a response, a RAG system can provide answers that are more closely aligned with the curriculum.
    
    This study explores the development and evaluation of a RAG-based academic support system designed specifically for computing students in programming modules. The system integrates an open-source LLM with a retrieval layer built from university-curated teaching materials. The aim of this research is to examine whether such a system can provide accurate, relevant, and reassuring academic support while reducing hesitation in seeking help.

    This study makes three contributions to research on educational AI and academic help-seeking. First, it provides a mixed-methods evaluation of a course-specific RAG system incorporating both student and academic perspectives, examining not only usability and perceived response quality but also trust, learning support, and help-seeking. Second, it examines course-specific RAG as a potential intermediary layer between independent study and formal academic support, exploring whether private, immediate, and module-aligned assistance can lower perceived barriers to initiating help-seeking. Third, it identifies design tensions of course-specific AI support, including the balance between accessibility and cognitive effort, curriculum alignment and conversational flexibility, and pedagogical scaffolding and learner autonomy. These findings provide insights into how course-specific generative AI might complement, rather than replace, existing academic support structures.
    
    The study is guided by the following research questions:

    \begin{itemize} 
        \item \textbf{RQ1:} How do students perceive the usefulness and relevance of a course-specific RAG system for academic support? 
        \item \textbf{RQ2:} To what extent do students perceive the system as reducing barriers associated with academic help-seeking? 
        \item \textbf{RQ3:} How do students evaluate the trustworthiness of responses generated by a system grounded in course materials?
        \item \textbf{RQ4:} How do academic staff perceive the system's value in reducing barriers to student help-seeking, and what tensions does this raise for maintaining pedagogical safeguards and scaffolded learning?
    \end{itemize}

    The remainder of the paper is structured as follows. Section~\ref{sec:lit_background} reviews the literature concerning help-seeking, generative AI, RAG, and trust in educational AI systems. Section~\ref{sec:sys_design} describes the design and technical implementation of Beacon. Section~\ref{sec:meth} outlines the research methodology including findings from an initial exploratory pilot study. Section~\ref{sec:results_discussion} then presents and discusses the main study findings. Sections~\ref{sec:limitations} and ~\ref{sec:future_work} consider the study's limitations and directions for future work, before Section~\ref{sec:conclusion} concludes the paper.

\section{Background and Literature}
    \label{sec:lit_background}

    Access to academic support is a fundamental component of student success in HE. However, access to support is not experienced equally. While universities provide a range of formal and informal support mechanisms, including lectures, tutorials, office hours, peer learning, and online resources, students differ considerably in their willingness and ability to engage with these opportunities. Factors such as confidence, anxiety, fear of judgement, prior educational experiences, and perceptions of belonging can all influence whether students seek help when encountering academic difficulties. Consequently, disparities in help-seeking behaviour may contribute to wider inequalities in learning experiences and academic outcomes, even when equivalent support provision exists.

    Recent advances in GenAI have introduced new possibilities for addressing these challenges. Students are increasingly turning to LLMs as an accessible source of immediate academic support, yet unrestricted AI systems frequently produce responses that are disconnected from institutional curricula, pedagogical intentions, or assessment expectations. This creates a tension between expanding access to learning support and ensuring that such support remains educationally appropriate, trustworthy, and equitable. This literature review is therefore structured as follows. First, the study context is established by exploring the difficulties in learning programming which can heighten help-seeking disparities. Second, factors influencing help-seeking behaviour in HE are examined before the growing role of AI in supporting learning are considered. The review concludes by proposing that course-specific Retrieval-Augmented Generation (RAG) systems offer a promising approach to reducing barriers to academic support while maintaining alignment with institutional teaching practices.

    \subsection{Learning Programming}
        
        The focal context for this study is students learning computer programming, a subject considered difficult to learn and where understanding can take years to develop \citep{robins2003learning, sentance2017computing}. The abstract nature of the subject means novice programmers often struggle, particularly when combining concepts or applying knowledge to new problems \citep{robins2003learning, saeli2011teaching}. This can create vicious cycles in knowledge development, especially when a typical course introduces new principles each session in an upward learning curve. However, a weak understanding of foundational concepts can inhibit understanding of more advanced principles \citep{duke2000teaching, saeli2011teaching}. 
        
        These issues are increased by diverse cohorts where some students enter HE with existing programming experience, while others do not. Sentance and Csizmadia \citeyear{sentance2017computing} argue skills disparities among students are one of the biggest challenges in teaching programming. Faced with peers they perceive as more competent, students can experience anxiety and reduced confidence, which impacts engagement \citep{morales2024connecting, rosenstein2020identifying, baker2019educational}. Furthermore, research shows that novice programmers struggle to seek help effectively and will either avoid help even when needed, or abuse help and bypass real learning \citep{marwan2020unproductive}.
        
    \subsection{Help-Seeking Behaviour in Higher Education}

        In HE more broadly students are expected to show independence in academic work, particularly with assessment tasks. To develop independence and achieve academic success, help-seeking behaviour is argued to play an important role \citep{fong2023academic}. However, students often avoid seeking support due to anxiety, fear of mistakes, and a lack of confidence \citep{Salim2022The}. In addition, a fear of seeming inadequate, the desire to appear competent, or being unaware of the support available can further inhibt help-seeking \citep{karabenick1991relationship}. This body of work has  been synthesised in a systematic review of $55$ studies spanning a decade of research, which identifies confidence, fear of judgement, and limited awareness of available support as recurring predictors of help-seeking avoidance across the higher education literature \citep{li2023college}.
        
        HE culture can inadvertently reinforce this avoidance by implying students should already possess skills to manage complex tasks, making help-seeking seem like a weakness \citep{sithaldeen2022student, robiullah2026navigating}. Within competitive environments, which may be caused by comparing oneself to peers, students may also internalise confusion, leading to heightened anxiety and poorer performance \citep{newton2017evidence}. Furthermore, large class sizes and limited access to academic staff can discourage students from asking for guidance \citep{tinto2012leaving}, while lecturers, who are often seen as a source of support, may themselves be uncertain about their role in student assistance \citep{Grayson1998Help-seeking}, or lack confidence in their ability to support diverse students \citep{Mcfarlane2016Tutoring}.

        Willingness to seek help is also influenced by students gender, age, cultural background and personality \citep{bornschlegl2020variables, ruihua2025understanding}. In particular, research has shown that mature students can face issues with imposter syndrome and confidence \citep{chapman2017using}. Therefore, as the diversity of the student population expands, so do the disparities regarding help-seeking behaviour. 

        Confidence influences help-seeking, with confident learners more likely to seek support \citep{Broadbent2023Help-seeking}. However, confident learners are also found to benefit the least from help-seeking strategies, whilst for unconfident learners help-seeking is positively associated with academic success \citep{Broadbent2023Help-seeking}. This creates a contradiction where those who need help the most are the ones who seek it the least despite the positive benefits it will offer \citep{ryan1998some, karabenick1991relationship, micari2021ok}. Again, this leads to educational disparities, with student confidence or self-efficacy influenced by prior experience \citep{hill2022evaluating} and personal or environmental factors \citep{miao2025influence} including race, age and gender \citep{huang2013gender,brown2021barriers, koh2022self, micari2021ok}. These issues are observed first hand by the authors, with some students being reluctant to seek clarification or help, even when struggling. This is then further evident in module feedback and assessment reflections where students comment that they should have made more contact with their tutor.

        Cultivating a supportive culture that normalises help-seeking is essential to address these issues and involves fostering belonging and trust \citep{sithaldeen2022student}, enhancing tutor training \citep{Mcfarlane2016Tutoring}, and promoting peer support \citep{Krisi2021The}. Additionally, LLMs have the potential to support academic learning by summarising information and providing feedback \citep{Laato2023AI-Assisted,radford2019language}. While these tools are beneficial, they can sometimes generate content that misaligns with course goals, leading students off-track \citep{Eager2023Prompting}. Carefully integrating LLMs, possibly through RAG frameworks, ensures course relevance by drawing from specific databases \citep{lewis2020retrieval}. Therefore, supporting students in overcoming help-seeking reluctance requires institutional efforts to normalise seeking assistance with value in exploring the possibilities enabled by artificial intelligence tools in supporting such efforts.

    \subsection{AI-Supported Learning in Higher Education}

        Artificial intelligence (AI) has become an increasingly prominent feature of HE, supporting activities including personalised learning, formative feedback, assessment, academic writing, and student support \citep{holmes2019ethics, luckin2016intelligence, selwyn2019should, gray2024bcj, gray2025mbcj}. Rather than functioning solely as administrative technologies, AI systems are increasingly positioned as learning companions capable of providing immediate explanations, answering questions, and supporting students outside scheduled teaching sessions.

        The emergence of LLMs and GenAI has accelerated this shift. Students now routinely use systems such as \textit{ChatGPT} to explain concepts, debug code, summarise readings, and clarify assessment requirements \citep{Laato2023AI-Assisted, Eager2023Prompting}. LLMs also help with language tasks, offering feedback on grammar and structure, as well as personalising learning content based on student needs \citep{radford2019language}. The availability of these systems provides students with immediate access to support regardless of time or location, potentially reducing reliance on traditional forms of academic assistance \citep{Laato2023AI-Assisted,yang2025strengths}. This flexibility may be particularly valuable for students who are reluctant to seek help from lecturers or peers due to anxiety, low confidence, or fear of judgement.

        While these systems offer new opportunities to support learning and teaching, they also introduce a number of challenges that affect their reliability in educational settings. Concerns have been raised about the accuracy of generated outputs, the presence of biases in training data, and the difficulty users may face when evaluating the quality of responses \citep{bommasani2021foundation, bender2021dangers}. 
        
        One widely discussed limitations of GenAI is their tendency to produce incorrect or fabricated information, sometimes referred to as “hallucinations”. Because these systems generate text by predicting likely word sequences rather than retrieving verified facts, they may produce outputs that appear convincing but contain inaccuracies. Alternatively, they may assume inappropriate levels of prior knowledge, or recommend approaches that differ from those expected within a particular module or assessment \citep{Eager2023Prompting,  Caines2023On,yang2025strengths}. This poses particular risks in educational contexts where students may rely on generated responses as authoritative sources of information \citep{ji2023survey}. A recent systematic review of GenAI in computer science education also found that hallucinated or misleading outputs can increase cognitive load during programming and debugging tasks. Consequently, this may disproportionately disadvantage students with limited prior programming experience or from under-resourced learning contexts, thereby widening rather than narrowing existing educational disparities \citep{adejumo2026systematic}.

        Furthermore, GenAI poses challenges related to transparency and accountability. The scale and complexity of modern language models make it difficult for users to understand how responses are produced or which sources influenced the output. As a result, educators and learners may find it difficult to assess the reliability of generated information or determine when AI tools should be trusted \citep{bommasani2021foundation}. 

        To address these concerns, students must critically engage with LLM-generated content, ensuring they use these tools to complement, rather than replace, their academic efforts \citep{Caines2023On, Yan2023Practical}. Without appropriate guidance, learners may struggle to critically evaluate AI-generated content.
        
        These opportunities and challenges suggest that educational AI should not simply aim to maximise automation or information access, but instead support meaningful learning within the context of institutional teaching practices. This has led increasing attention towards AI systems that can provide contextually grounded, course-specific guidance while remaining aligned with approved educational resources.

    \subsection{Retrieval-Augmented Generation for Educational Support}
    
\begin{figure*}[t]
            \centering
            \includegraphics[width=\linewidth]{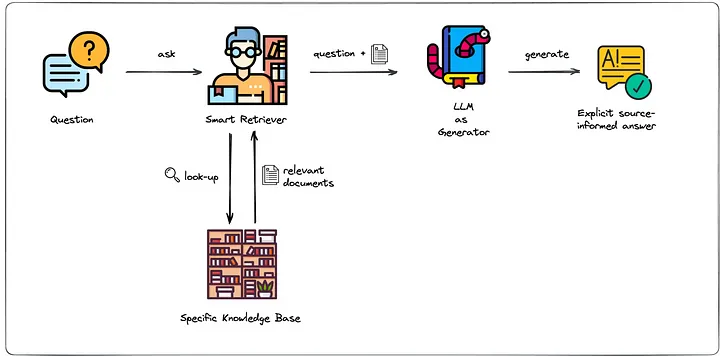}
            \caption{A visual demonstration of how RAG can help inform the LLMs with relevant material stored within a vector database \citep{greer2024rag_figure}. The user will ask a question, which, in turn, the smart retriever, a vector similarity searcher, will query the Vector database to retrieve the relevant information to present to the LLM as a generator. The LLM then produces an output representing the user's question and the additional information retrieved from the vector database. In our context, lesson content and assessment details are stored in the vector database to aid the students.}
            \label{fig:LLM_visualisation}
        \end{figure*}

        RAG combines information retrieval with LLM generation. Rather than rely solely on the internal knowledge of a LLM, RAG systems retrieve relevant documents from a predefined knowledge base and use them to inform the generated response (see Figure \ref{fig:LLM_visualisation} for a visual representation).
        
        This approach offers particular promise for educational applications because it allows responses to be grounded in course materials, institutional resources, or other curated sources. As a result, students receive answers that are contextually relevant and aligned with the curriculum.
        
        RAG improves LLMs by allowing them to access external databases, ensuring they provide up-to-date accurate information. Traditional models like GPT rely only on the information they were trained on, which can become outdated \citep{brown2020language}. RAG overcomes this by combining the ability of the model to generate text with real-time information retrieval, leading to more accurate and relevant responses \citep{lewis2020retrieval}.
    
        In HE, RAG can help students retrieve relevant academic papers, articles, and reliable sources, making research easier. It can also generate summaries or explanations, helping students engage with large amounts of material for assignments and exams \citep{izacard2020leveraging}. Furthermore, RAG can personalise learning by providing information tailored to a student's needs, ensuring they receive up-to-date content \citep{borgeaud2022improving}.

        The application of RAG specifically to course- and institution-specific student support has grown rapidly. A recent survey of $47$ studies on RAG chatbots in education maps this emerging sub-field across application type, knowledge domain, and evaluation approach \citep{swacha2025rag}. Within computing education specifically, several course-specific implementations have been reported: \citet{neumann2025llm} developed and evaluated an LLM-driven RAG chatbot for a databases and information systems course, reporting an 88\% accuracy rate against course content; \citet{lang2025aipowered} evaluated a RAG chatbot in an online programming course, finding that students with greater prior knowledge engaged more with advanced queries; \citet{alsafari2024towards} compared intent-based and RAG-based teaching assistants for a data mining course; \citet{nemeth2025exploring} piloted a RAG-based tutor across four courses at two universities, reporting expert-coded accuracy rates alongside student and lecturer perceptions; and \citet{tranhuuvan2026course} reported a course-specific agentic RAG chatbot for IT student support with a comparable architecture and privacy-aware design. Beacon extends this emerging body of work by combining both student and academic-staff evaluation within a single study, and by considering  not only usability and response accuracy but also the impact on trust, learning support, and help-seeking.
    
    \subsection{Trust, Transparency and Human-Centred Educational AI}

        Trust is widely recognised as a prerequisite for the effective adoption of AI in education \citep{DfE2023GenAI, QAA2023AI, Jisc2023GenAI}. Students must feel confident that AI-generated guidance is reliable, relevant, and appropriate to their learning context, while educators require assurance that AI systems support rather than undermine established pedagogical practices \citep{gonsalves2025clear, gray2025rendering, holmes2019ethics, nazaretsky2022trust}. For educational AI systems intended to support learning, trust influences not only whether students choose to use the system, but also whether they engage critically with the guidance provided and integrate it appropriately into their learning.

        Research suggests that trust is shaped by more than technical performance. Transparency, explainability, and opportunities for human oversight contribute to how users perceive the reliability and credibility of AI-generated responses \citep{glikson2020human, khosravi2022explainable}. Systems that provide little insight into how responses are generated may be perceived as opaque "black boxes", making it difficult for students and educators to judge the validity or appropriateness of the information provided. Explainable AI (xAI) therefore seeks to improve transparency by enabling users to understand the basis of AI-generated outputs through meaningful explanations or supporting evidence \citep{adadi2018peeking}. Within educational contexts, this transparency can help students verify AI-generated guidance against trusted learning resources while enabling educators to evaluate whether responses remain aligned with module learning outcomes and assessment expectations.
        
        However, transparency alone does not automatically establish trust. Students must also perceive AI systems as intuitive, accessible, and responsive to their learning needs. Consequently, human-centred design has become an increasingly important principle within educational AI, emphasising that systems should be designed around the needs, experiences, and educational contexts of their intended users rather than technological capability alone \citep{dix2016human, shneiderman2022human, becker2020learn}. Human-centred approaches advocate iterative design, user involvement, accessibility, and continuous refinement through evaluation with end users. Such approaches recognise that technically accurate responses are insufficient if learners find systems difficult to use, cannot determine when responses should be trusted, or are unable to relate AI-generated guidance to their existing learning resources.
        
        For course-specific AI systems, these principles suggest that successful educational support depends not only on retrieving relevant information, but also on presenting that information in ways that encourage confidence, independent learning, and appropriate help-seeking. Interfaces should therefore minimise barriers to use, communicate the provenance of generated responses, and enable students to verify explanations against approved teaching materials before applying them to their own work. Such design choices encourage students to remain active participants in the learning process rather than passive recipients of AI-generated answers \citep{jin2023supporting, lan2025qualitative}.
        
        Taken together, the literature suggests that trustworthy educational AI emerges from the combination of technically grounded responses, transparent presentation of supporting evidence, and human-centred design. Rather than viewing AI solely as a means of increasing efficiency, these principles position educational AI as a mechanism for widening equitable access to academically appropriate learning support. This perspective is particularly relevant for students who may be reluctant to seek help through conventional channels due to anxiety, low confidence, or fear of judgement. These principles therefore informed both the design of Beacon and the methodological approach adopted in its evaluation, providing the foundation for a course-specific AI system intended to complement existing teaching while reducing barriers to academic support.

\section{System Design}
    \label{sec:sys_design}

    The student support tool, referred to throughout this paper as Beacon,\footnote{Beacon was developed and evaluated under the working name DebuggyDuck during its initial pilot development. It was renamed prior to the main study reported here; both names refer to the same underlying system.} was designed to provide immediate, course-specific academic guidance that complemented existing teaching while reducing barriers to help-seeking. Rather than relying on unrestricted LLMs, which may generate responses beyond the scope of a module, Beacon grounds responses exclusively in approved teaching materials. This section describes the educational design objectives, the RAG architecture adopted to achieve them, the construction of the course-specific knowledge base, and the user interface developed through iterative evaluation.

        \subsection{Design Objectives}

            The primary objective of Beacon was to provide students with immediate, course-specific academic guidance that complemented existing teaching while reducing barriers to help-seeking. Rather than replacing lecturers, tutorials, or existing learning resources, Beacon was designed to act as an accessible first point of support when students encountered difficulties outside scheduled teaching. This was motivated by the recognition that many students are reluctant to seek academic help due to factors such as anxiety, fear of judgement, uncertainty regarding module expectations, or lack of confidence in their own understanding. Thus, a core aim for Beacon was to lower the threshold for accessing support while encouraging students to engage more confidently with formal teaching provision when required.

            Furthermore, the authors observed a sharp-rise in academic misconduct cases involving generative AI in recent years stemming from students consulting unrestricted tools such as \textit{ChatGPT} for programming assistance and receiving responses that draw on techniques, libraries, or approaches beyond the scope of the module or assessment. When such content is incorporated into submitted work, the resulting mismatch with taught material and student understanding can prompt closer scrutiny from module staff, and in some cases leads to formal academic misconduct proceedings. Because Beacon's responses would be grounded exclusively in approved teaching materials, the aim is to reduce this particular pathway to misconduct by keeping guidance within the scope of the module. 
            
            To achieve these aims Beacon was developed around four educational design objectives:
            
            \begin{enumerate}
                \item \textbf{Provide course-specific academic support:} Responses should remain aligned with the module learning outcomes, teaching materials, and assessment expectations rather than relying solely on the broad knowledge of a general-purpose LLM.
            
                \item \textbf{Support independent learning:} Beacon should encourage students to understand concepts and develop problem-solving skills through explanations and guidance rather than simply providing complete solutions.
            
                \item \textbf{Promote transparency and trust:} Students should be able to understand where information originates enabling them to verify explanations against official module resources. For example, by stating the material was from session presentation $x$ and slide $y$, or from handout $z$.
            
                \item \textbf{Complement existing teaching practices:} Beacon should enhance, rather than replace, interactions with lecturers and tutors by providing timely support between teaching sessions and encouraging students to seek further assistance when appropriate.
            \end{enumerate}
            
            These objectives informed each stage of Beacon's development, from the construction of the course-specific knowledge base and retrieval pipeline to the design of the user interface. Rather than viewing the LLM as the primary source of knowledge, Beacon treats approved teaching materials as the authoritative foundation for all responses. RAG was therefore adopted not simply as a technical solution, but as an educational design choice that enables AI-generated support to remain transparent, contextually relevant, and aligned with institutional teaching practices.

    \subsection{System Architecture}
        \label{subsec:sys_arch}

        The student support tool adopts a modular architecture that combines a course-specific knowledge base with a LLM to provide contextually grounded responses. Rather than relying solely on the pretrained knowledge of the LLM, every response is generated using relevant teaching materials retrieved from the module knowledge base. This ensures that generated guidance remains aligned with the learning outcomes, teaching content, and assessment expectations of the module while preserving the conversational interaction expected of modern AI assistants.

        The overall workflow is illustrated in Figure~\ref{fig:LLM_visualisation}. Teaching materials are first processed and indexed within a vector database to create a searchable knowledge base. When a student submits a question, the query is converted into a semantic representation and matched against the indexed course materials to identify the most relevant content. The retrieved passages are then incorporated into the prompt supplied to the LLM, allowing responses to be generated using information drawn directly from approved module resources rather than relying exclusively on the model's internal knowledge.

        This architectural approach was selected to support the educational design objectives outlined in the previous section. Grounding responses in module-specific resources helps ensure that explanations remain consistent with the material taught during class, reducing the likelihood that students receive guidance beyond the intended scope of the module. Furthermore, because responses are explicitly linked to retrieved teaching materials, students are encouraged to verify explanations against the original resources, supporting transparency, independent learning, and trust in the system.

        The architecture comprises four principal components: 
        \begin{enumerate}
            \item a curated knowledge base containing approved teaching resources;
            \item a semantic retrieval component responsible for identifying relevant learning materials;
            \item a LLM responsible for generating conversational responses; and
            \item a web-based user interface through which students interact with the system.
        \end{enumerate}
        
        Therefore, ultimately the LLM functions as the conversational interface, whereas approved teaching materials remain the authoritative source of academic content.
        
        The following sections describe each of these components in greater detail.

        \subsubsection{Knowledge Base Construction}
            \label{subsec:knowledge_base}

            The knowledge base was constructed from teaching materials approved for use within the participating computing modules. These materials included lecture slides, laboratory exercises, formative assessment activities, module handbooks, assessment guidance, and supplementary learning resources. Including multiple resources enabled Beacon to retrieve both conceptual explanations and practical guidance while remaining aligned with the content and expectations communicated through formal teaching.
            
            To provide a consistent representation across different document types, the teaching materials were prepared in Markdown. Presentation materials were authored using MARP, while supporting documents were stored as standard Markdown files. This format preserved meaningful structural features, including headings, lists, code blocks, and links, while removing visual and presentation-specific formatting that was not required for semantic retrieval.
            
            Before indexing, each document was segmented into smaller textual chunks. This approach was intended to retain sufficient contextual information within each segment while preventing large documents from reducing retrieval precision.
            
            Each chunk was stored together with metadata describing its original source. Depending on the resource type, this metadata included the module or session identifier, document title, resource type, topic, slide number, and the original text. Retaining this information enabled retrieved passages to be traced back to their source materials and displayed alongside generated responses. This supported transparency by allowing students to verify AI-generated explanations against the official module resources.
            
            A vector embedding was generated for each chunk using an embedding model. The embeddings were stored in a PostgreSQL database using the pgvector extension. At retrieval time, the embedding of the student's query was compared with the stored document embeddings. The TOP-$7$ highest-ranked chunks were returned, with no fixed similarity threshold, and were subsequently incorporated into the generation prompt.
            
            The knowledge base was indexed before deployment and updated whenever approved teaching materials were added or revised. Only materials selected by the module teaching team were included, ensuring that the retrieved content remained course-specific and pedagogically appropriate.

        \subsubsection{Retrieval-Augmentation Generation Pipeline}
            \label{subsec:RAG_pipeline}

\begin{figure*}[t]
                    \centering
                    
                    \begin{tikzpicture}[
                        node distance=0.8cm and 1.6cm,
                        every node/.style={font=\small},
                        process/.style={
                            rectangle,
                            rounded corners,
                            draw,
                            align=center,
                            minimum width=3.6cm,
                            minimum height=0.8cm
                        },
                        decision/.style={
                            diamond,
                            draw,
                            aspect=2,
                            align=center,
                            inner sep=1pt
                        },
                        line/.style={->, thick}
                        ]
                        
                        \node[process] (input) {Student Question};
                        
                        \node[decision, below=of input] (decision) {Initial\\Question?};
                        
                        \node[process, below left=1.5cm and 2cm of decision] (embed)
                        {Generate Query Embedding};
                        
                        \node[process, below right=1.5cm and 2cm of decision] (rewrite)
                        {Rewrite Query\\using Conversation Context};
                        
                        \node[process, below=2cm of decision] (retrieve)
                        {Semantic Similarity Search\\Vector Database};
                        
                        \node[process, below=of retrieve] (chunks)
                        {Retrieve Top-$k$\\Document Chunks};
                        
                        \node[process, below=of chunks] (prompt)
                        {Construct LLM Prompt\\
                        \small
                        System Prompt\\
                        + User Question\\
                        + Retrieved Content\\
                        + Conversation History};
                        
                        \node[process, below=of prompt] (llm)
                        {Large Language Model};
                        
                        \node[process, below=of llm] (response)
                        {Grounded Response\\Returned to Student};
                        
                        \draw[line] (input) -- (decision);
                        
                        \draw[line] (decision) -- node[left]{Yes} (embed);
                        \draw[line] (decision) -- node[right]{No} (rewrite);
                        
                        \draw[line] (embed) |- (retrieve);
                        \draw[line] (rewrite) |- (retrieve);
                        
                        \draw[line] (retrieve) -- (chunks);
                        \draw[line] (chunks) -- (prompt);
                        \draw[line] (prompt) -- (llm);
                        \draw[line] (llm) -- (response);
                    
                    \end{tikzpicture}
                    
                    \caption{Retrieval pipeline used to generate grounded responses. Initial questions are embedded directly for semantic retrieval, whereas follow-up questions are first rewritten into self-contained queries using the preceding conversation before retrieval is performed. The final prompt supplied to the language model combines the system prompt, user query, retrieved teaching materials, and conversation history.}
                    \label{fig:retrieval_pipeline}
                    
                    \end{figure*}
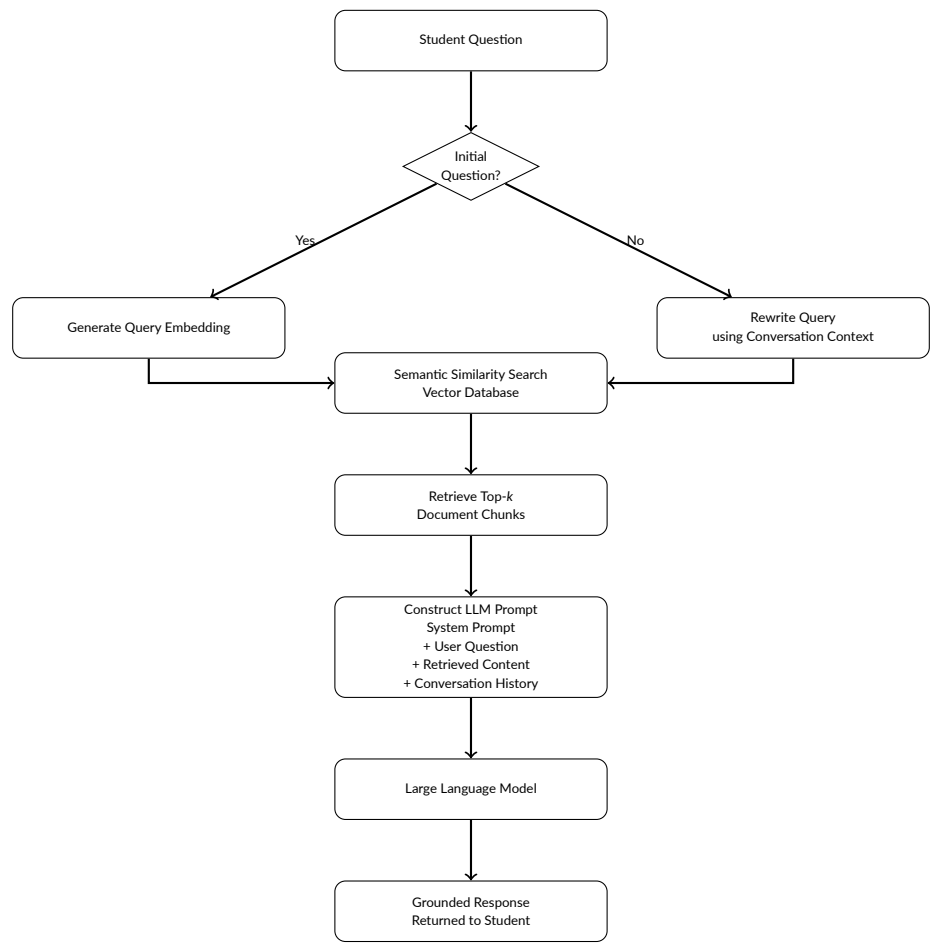

                The retrieval pipeline (Figure~\ref{fig:retrieval_pipeline}) distinguished between the initial user message and subsequent conversational turns. When a student submitted the initial question, the message was converted directly into an embedding and used to perform a semantic similarity search over the indexed course materials.

                For each subsequent message, the system generated a self-contained retrieval query using the latest user message and the preceding conversation history. This step was introduced because conversational follow-ups may be semantically underspecified when considered in isolation. For example, a request such as \textit{``Can you explain more?''} does not identify the concept to which it refers and would therefore be unlikely to retrieve relevant material if embedded directly. Rewriting the message using the preceding context produced a more informative search query while preserving the student's intended meaning.

                The initial or rewritten query was embedded and compared with the knowledge-base embeddings using cosine similarity. The TOP-$7$ highest-ranked document chunks were retrieved and incorporated into the final generation prompt. This prompt comprised the system instructions, the student's original message, the retrieved course materials, and the relevant conversation history.

                The rewritten query was used for retrieval purposes only. The student's original message was retained within the final prompt so that the generated response remained natural and appropriate to the conversation. This process enabled multi-turn interactions to remain grounded in relevant course content, even where the student's latest message contained little contextual information.

        \subsubsection{Prompt Construction}
            \label{subsec:prompts}

            The retrieved document segments were incorporated into a structured prompt together with the student's question before being supplied to the language model. The prompt instructed the model to answer using only the retrieved teaching materials, avoid speculation when insufficient evidence was available, and encourage conceptual understanding rather than directly completing assessed work. Where relevant, the generated response was accompanied by references to the original teaching resources, allowing students to verify explanations against official module content.

    \subsection{User Interface}
        \label{subsec:UI}
        
        Students interacted with Beacon through a conversational interface that enabled them to submit questions in natural language and receive contextually grounded responses. The application is designed to resemble familiar chat interfaces to reduce barriers to adoption and encourage student engagement. Responses were presented in a supportive, conversational tone to encourage exploration and independent learning.
        
\begin{figure}
            \centering
            \includegraphics[width=\linewidth]{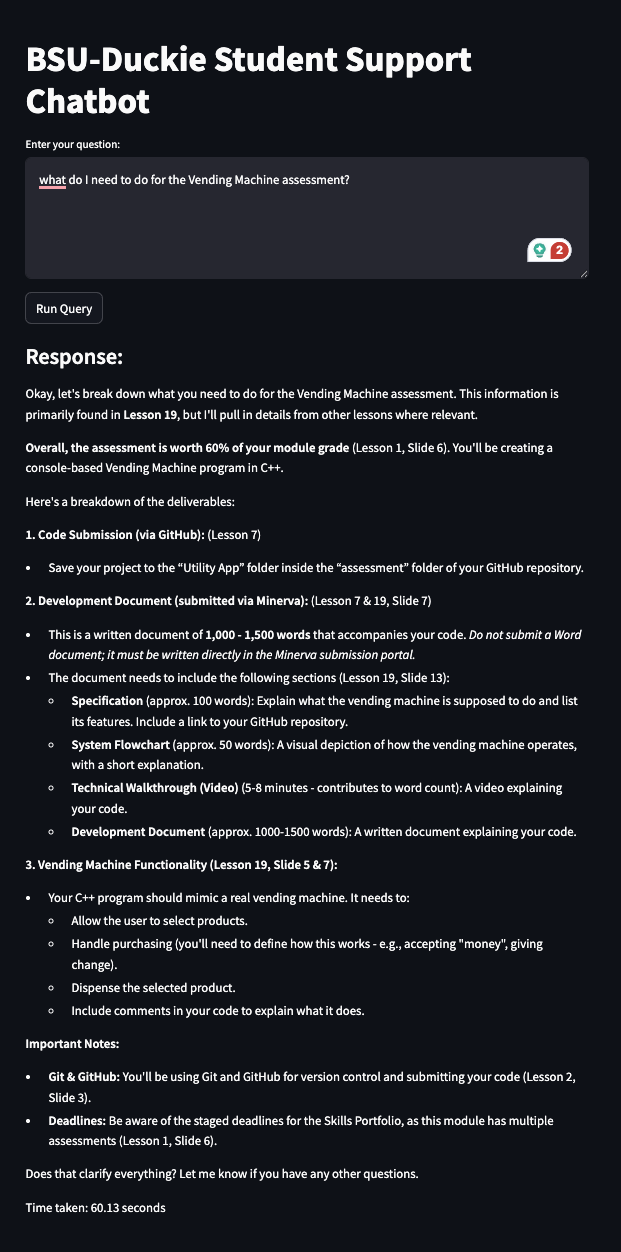}
            \caption{The Version~1 web application interface. The pilot implementation displays the generated response together with the supporting source material used to produce the answer.}
            \label{fig:RAG-app}
        \end{figure}

        Although versions of Beacon developed for the exploratory pilot and main study share the same underlying RAG architecture, the implementation evolved throughout the project. The pilot application (Version~1) was developed using Streamlit to enable rapid prototyping and iterative testing with students. The pilot interface adopted a deliberately minimalist design, consisting of a single text input field and submission button. After a question was submitted, the generated response was streamed beneath the input area along with the relevant supporting content (Figure~\ref{fig:RAG-app}). This lightweight interface prioritised rapid development and evaluation over visual sophistication, allowing the focus of the pilot study to remain on the effectiveness of the underlying RAG pipeline.
        
\begin{figure*}
            \centering
            \includegraphics[width=\linewidth]{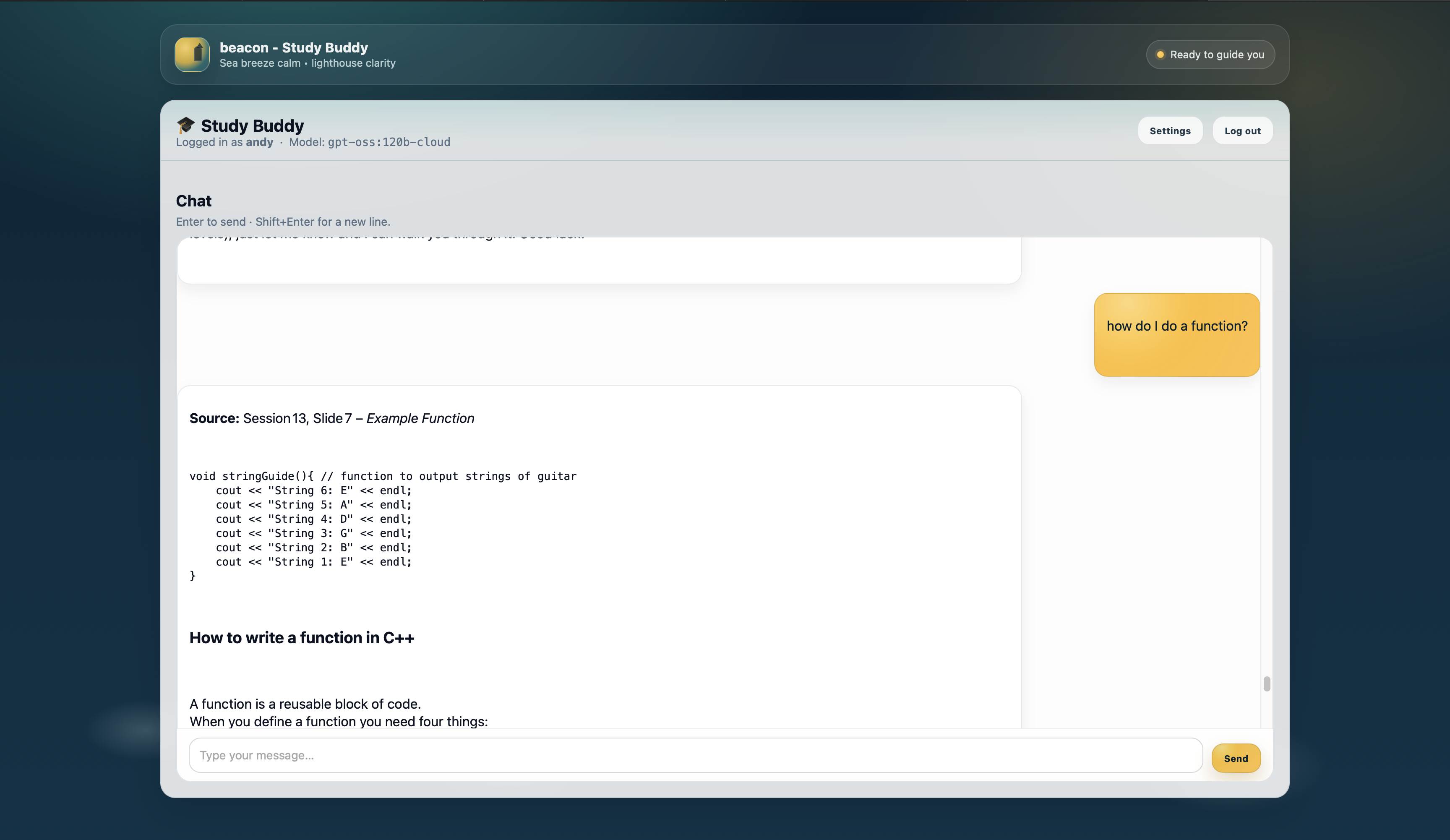}
            \caption{The Version~2 web application interface. The redesigned chat interface presents generated responses alongside the supporting source material used to produce each answer.}
            \label{fig:RAG-app-v2}
        \end{figure*}

        Following the pilot study, the application was reimplemented using Flask to provide a more scalable architecture suitable for web hosting and wider deployment. Feedback obtained during the pilot  (see Section~\ref{sec:pilot}) also informed several interface refinements based around a more conventional chat interface that more closely reflected commercial conversational AI systems. This provided a more intuitive interaction model while preserving transparency by continuing to display the supporting source material alongside each response (Figure~\ref{fig:RAG-app-v2} ). In addition, the underlying language model was upgraded from ``GPT-OSS-20B'' to ``GPT-OSS-120B'' in Version~2. Consequently, the primary differences between the two versions were improvements to the user interface, deployment architecture, and language model capability, rather than changes to the core RAG workflow.

    \subsection{Implementation Details}
        \label{subsec:Implementation}

        The application was implemented using Python's Flask library and deployed as a web-based application hosted on Python Anywhere. The knowledge base was stored using PostgreSQL with the pgvector extension to support semantic similarity search. Teaching resources were embedded using Nomic Embed Text and indexed within the vector database prior to deployment.
        
        The complete application was developed using open-source technologies, allowing Beacon to be deployed locally or within institutional infrastructure without dependence on commercial AI platforms. This supports institutional control over teaching resources, model selection, and student data while facilitating future adaptation to different modules or programmes.

\section{Methodology}
    \label{sec:meth}

    \subsection{Research Design}
    
        This study adopted a Design-based research (DBR) approach, which aims to improve educational practice through iterative design in authentic settings \citep{wang2005design}. A typical DBR process follows five steps:

        \begin{enumerate}
            \item Identify a problem (established in Section~\ref{sec:intro} and Section~\ref{sec:lit_background})
            \item Design an intervention (outlined in Section~\ref{sec:sys_design})
            \item Implement and evaluate (see Section~\ref{sec:pilot} and Section~\ref{sec:results_discussion})
            \item Refine the design (as illustrated by the changes between the pilot and main study)
            \item Develop theory and/or design principles (see Section~\ref{sec:results_discussion} and Section~\ref{sec:conclusion})  
        \end{enumerate}
        
        Using the DBR approach the research followed an iterative design process comprising two stages. An initial pilot study evaluated a first prototype, focusing on the technical feasibility of the RAG approach and the usability of the interface. Findings from this pilot informed the development of a second version of Beacon, incorporating refinements to both the user interface and underlying infrastructure. This revised system was subsequently evaluated through a larger user study. This iterative cycle aligns with the DBR approach where successive refinement based on user feedback enables systems to better address learner needs \citep{hoadley2022design}.
        
        DBR uses various approaches to evaluate the effectiveness of the designed intervention. This combination of methods is argued to increase the validity and applicability of the research \citep{wang2005design}. As such, data collection combined both qualitative and quantitative data through structured questionnaires and semi-structured interviews. Questionnaires provided quantitative measures of students' perceptions of Beacon, while interviews explored participants' experiences in greater depth, allowing themes to emerge regarding usability, trust, learning behaviours, and opportunities for improvement. Alongside student feedback, interviews were also conducted with lecturers to gain their perspectives of Beacon. The combination of methods and gathering of multiple perspectives enabled the triangulation of findings, providing both breadth and depth when evaluating the educational value of Beacon \citep{cohen2009research}. 

    \subsection{Exploratory Pilot Study}
        \label{sec:pilot}
    
        In line with the iterative nature of the DBR approach, an exploratory pilot study was used to assess the feasibility and proof-of-concept validity of Beacon. During the pilot, Beacon was not deployed for independent student use; instead, it was presented through a structured demonstration that illustrated its functionality and intended use. The demonstration included examples of typical student queries, such as requests for clarification of programming concepts and interpretation of assignment requirements. This approach enabled participants to observe how the system retrieved relevant information and generated responses grounded in course materials.
            
        The pilot prioritised demonstrating core functionality and pedagogical alignment rather than full technical maturity. This allowed for an initial evaluation of how students perceived the system before further development and refinement.
    
        \subsubsection{Pilot Study Procedure}
        
            The pilot study was conducted with undergraduate computing students who attended scheduled teaching sessions where Beacon was demonstrated. Multiple groups of students participated in these sessions, providing a cross-section of the cohort within the module context.
            
            During each session, Beacon was introduced and demonstrated by the researcher. Students were shown how queries could be submitted and how the system generated responses using module-specific materials. The demonstration included a series of representative examples designed to reflect realistic academic scenarios, allowing students to understand how the system might be used to support their learning.
            
            Following the demonstration, students were invited to complete a structured questionnaire designed to capture their perceptions of the system. The questionnaire focused on students’ impressions of the system’s clarity, usefulness, relevance to course content, and perceived trustworthiness. It also explored how students anticipated using such a system in relation to their existing study practices.
                
            Participation in the study was voluntary, and all responses were collected anonymously. The aim of this procedure was to capture students’ immediate perceptions following exposure to the system, rather than to evaluate long-term patterns of use. A total of nine valid responses were collected and used for subsequent analysis.
    
        \subsubsection{Pilot Study Evaluation}
        
            Findings from the pilot indicated a generally positive student response to the system, alongside identifiable areas for improvement.
            
            Regarding clarity, 44\% of respondents reported that the system was very clear, with a further 44\% indicating it was somewhat clear. A small proportion remained neutral, suggesting that additional explanation or demonstration may improve accessibility. Confidence in response quality was also high, with 56\% of participants reporting that outputs were consistently logical and the remainder indicating that responses were mostly reliable.
    
            Perceived relevance was particularly strong, with 89\% of participants rating the system’s responses as highly aligned with course materials. This suggests that the retrieval mechanism successfully grounded responses in module-specific content, addressing a common limitation of general-purpose AI tools.
            
            Students also reported that the system was useful in supporting assessment understanding, with all participants indicating that it was either definitely or somewhat helpful. This reinforces the system’s potential role as a first point of support when students encounter difficulty.
    
            Trust in the system was more cautious. While the majority of participants expressed some level of trust, only a small proportion indicated complete confidence in the responses. This suggests that, although the system is perceived as useful, students do not treat it as an authoritative source. Instead, it is positioned as a supplementary tool, with many indicating that they would still verify important information.
            
            Participants also recognised potential risks associated with the system. Concerns were raised regarding the possibility of incorrect information and the risk of over-reliance, particularly if students use the system as a substitute for independent thinking. These concerns highlight the importance of transparency and appropriate integration into teaching practice.
            
            Despite these limitations, many participants indicated that the system could reduce hesitation in seeking help, particularly by enabling private interaction. This suggests that AI-based support systems may play a role in addressing barriers associated with traditional help-seeking behaviours.
            
            Overall, the findings suggest that a course-specific RAG system can provide relevant and accessible academic support, while also introducing important considerations relating to trust, transparency, and student dependency. The feedback from the pilot study resulted in enhancements to the user interface, deployment architecture, and language model capability as described in Section ~\ref{sec:sys_design}.

    \subsection{Participants}

        Participants in the main study were undergraduate students enrolled on computing-related degrees at a British University. Students were invited to participate voluntarily in the evaluation of Beacon as an AI-supported learning tool. Before participation, students were informed of the purpose of the study and advised responses would be collected anonymously and used solely for research purposes. Participating students received a £10 gift voucher, redeemable at a retailer of their choice, as compensation for their time.
        
        A total of fifteen students participated in the main study, a figure widely cited as sufficient to identify all critical usability problems \citep{faulkner2003beyond, nielsen2000fiveusers}. Participants represented a range of computing disciplines, including Computing, Cyber Security, Games Development, and Creative Computing to ensure the evaluation reflected diverse programming backgrounds, varying levels of confidence, and differing degrees of familiarity with generative AI technologies. While the majority of participants were first-year students, second- and third-year students were also represented, enabling perspectives to be gathered from learners at different stages of study.
        
        Before interacting with Beacon all participants completed a pre-experiment questionnaire to establish baseline information relating to programming confidence, help-seeking behaviours, use of existing module resources, and prior engagement with generative AI systems. 

        Interaction with Beacon involved completing six task scenarios, a method recommended in usability testing for ensuring users meaningfully engage with the task \citep{mccloskey2014task}. These scenarios were design to reflect real-world usage and included questions and issues commonly experienced programming lessons (See Appendix ~\ref{app:student_tasks}). Nine students completed the tasks during a two-hour in person workshop with the remaining six completing the tasks remotely.  
        
        After completing the task scenarios participating students completed a post-experiment questionnaire evaluating the system's usability, clarity, relevance, trustworthiness, learning support, and potential influence on future help-seeking behaviours. Although responses were collected anonymously and no personally identifiable information was recorded, each participant used a unique anonymous identifier. This enabled pre- and post-experiment questionnaire responses to be linked at the individual level while preserving participant anonymity. Five of the fifteen students subsequently took part in semi-structured interviews to explore their experiences in greater depth (receiving a further £10 voucher for their time).

        To complement the student evaluation, four academic staff independently reviewed Beacon from a pedagogical perspective. All reviewers had experience teaching on the module associated with the study and represented a range of disciplinary backgrounds, including Computing, Cyber Security, Creative Computing, and Games Development. Collecting perspectives from both students and academics enabled Beacon to be evaluated from both learner and educator viewpoints.
        
        Ethical approval for the study was obtained through the University's institutional ethics process, and no personally identifiable information was collected throughout the study.

    \subsection{Data Collection}
    
        \subsubsection{Questionnaires}

            Two structured questionnaires were used during the evaluation. The pre-experiment questionnaire collected demographic information together with measures relating to programming confidence, existing support strategies, and prior experience of generative AI technologies. The post-experiment questionnaire evaluated students' perceptions of Beacon following use, including measures of usability, usefulness, clarity, trustworthiness, relevance, transparency, and perceived support for independent learning. The questionnaires included five-point agreement items, frequency-rating items, comparative-rating items, supplemented by optional open-ended responses to capture additional comments.
    
        \subsubsection{Semi-Structured Interviews}
        
            Semi-structured interviews were used to gain greater insight into participant experiences of using Beacon. Semi-structured interviews occupy a methodological position between structured questionaries and open ethnographic observation, allowing researchers to investigate predefined topics while retaining the flexibility to explore unexpected issues raised by participants \citep{blandford2013eliciting,blandford2008evaluating}. Thus, this approach sought to understand not only participants' opinions of Beacon but also the reasoning behind those opinions.
            
            A total of nine interviews were conducted. Five with students and four with academic staff. Collecting perspectives from both students and academics enabled Beacon to be evaluated from both learner and educator viewpoints.
            
            Students were randomly selected for the interviews, with no selection criteria applied other than their indication of willingness to participate on the study consent form. Interview questions focused on perceptions of usability, trust, learning support, transparency, and potential improvements. Follow-up questions enabled participants to elaborate on their experiences and provide examples of how Beacon influenced their learning.
            
            Interviews with academic staff focused on the educational appropriateness of the generated responses, alignment with module materials, and the potential integration of Beacon into existing teaching and student support practices. 
            
    \subsection{Data Analysis}

        Questionnaire responses to quantitative measures were analysed to calculate response frequencies and percentages, enabling comparison of student perceptions towards Beacon and its potential impact on learning and help-seeking behaviour. Responses to qualitative free-text questions were analysed analysed using a thematic approach to identify recurring insights and emergent themes. Data analysis was supported by Microsoft Copilot, which was used to assist in summarising quantitative results and identifying potential themes within the qualitative data. All analyses and AI-assisted outputs were reviewed and validated by the authors to ensure accuracy, consistency, and appropriate interpretation of the findings.

        Interviews were conducted via Microsoft Teams, with transcripts generated automatically. Immediately after the interview the researcher reviewed the transcripts to verify accuracy, correct errors and remove identifying information. The final transcripts were analysed with the support of Microsoft Copilot using the six-step process for AI-assisted thematic analysis described by \citep{naeem2025thematic}. The use of AI in thematic analysis is argued to be more time-efficient, reduce human bias and uncover insights that might otherwise be overlooked. As a result AI can enhance and improve the efficiency of the analysis \citep{naeem2025thematic}. To ensure accuracy, outputs generated by Copilot were reviewed and verified by the researchers at each stage of the analysis. Consistent with recommendations for qualitative research, careful documentation of the six-step process, including all prompts used to generate findings, have been retained to enhance the credibility and trustworthiness of the findings \citep{blandford2002case,kamsin2012personal,furniss2011confessions}.

\section{Results \& Discussion}
    \label{sec:results_discussion}

    \subsection{Student Views}

        \subsubsection{Pre-Experiment Questionnaire}
            
            The pre-experiment questionnaire explored students’ existing help-seeking behaviours, confidence levels, and use of AI-based support tools prior to interacting with Beacon. Overall, students reported moderate levels of programming confidence, with a mean self-rated programming ability of approximately $6.25$ out of $10$. The grades achieved by students in the foundational programming module ranged from third-class to first-class honours, indicating a diverse range of academic attainment among participants.
    
            The findings revealed an interesting tension between students’ willingness to seek support and their preference for independent problem-solving. Although participants generally reported feeling comfortable asking lecturers for help, many also demonstrated signs of hesitation and avoidance when encountering difficulties. The majority of students (62.5\%) indicated they sometimes avoided asking for help even when they needed it, while a greater proportion  reported anxiety when they did not understand a topic (75\%). At the same time, there was a preference for autonomy, with most students agreeing that they preferred to attempt solving problems independently before seeking support (87.6\%). These findings suggest that, while institutional support structures are available, many students still attempt to manage uncertainty privately before engaging with formal sources of assistance.
    
            Student responses also highlighted widespread use of GenAI tools within existing study practices. The majority of participants reported using GenAI either sometimes or often when working on programming-related tasks, while only a smaller subset (25\%) indicated that they never used such systems. AI tools were used across a broad range of activities, including debugging, concept explanation, assignment clarification, code interpretation, study planning, and academic writing support. Several students described AI systems as particularly valuable because they offered rapid responses, adaptive explanations, and personalised support that could be tailored to individual levels of understanding. Others highlighted that AI systems allowed them to ask follow-up questions without feeling judged or pressured, making them easier to engage with than traditional support channels in some situations.

            \begin{quote} \textit{"The use of AI tools allows for personalised responses and follow up questions, which is much more flexible and efficient than the module-provided information and third party websites."} Participant melo-f6 \end{quote}
            
            Despite this widespread use of AI, students did not position these tools as entirely replacing traditional module resources. Participants continued to identify module-provided content, such as lecture slides, recordings, GitHub repositories, and workshop materials, as the most helpful forms of support because these resources were directly aligned with course expectations and assessment requirements. Students emphasised the importance of contextual relevance, noting that general online resources or unrestricted AI systems could sometimes provide explanations that were too advanced, insufficiently tailored to their level of study, or disconnected from the approaches used within the module itself.

            \begin{quote} \textit{"Everything there is more direct to the level of coding that is expected of us, AI and even youtube can often give you things we haven't yet been taught or don't understand."} Participant bano-k7 \end{quote}
    
            The qualitative responses further suggested that many students engage with AI critically and iteratively, rather than relying passively on generated outputs. Several participants described refining prompts, requesting simplified explanations, verifying responses independently, and using AI systems primarily as a mechanism for clarification rather than direct answer generation. 

            \begin{quote} \textit{"AI tools allow clarification using online materials. In the event I do not understand what generative AI outputs, I can ask it specific relevant questions such as what a specific sentence meant in the context, where it can clarify in detail for me."} Participant feno-j5 \end{quote}
            
            However, concerns surrounding trust and reliability were also evident throughout the dataset. Students acknowledged that AI systems could sometimes provide incorrect or misleading information and recognised the risks associated with over-reliance on generated responses. In particular, some participants noted that AI explanations occasionally exceeded their current level of understanding, which could further contribute to confusion rather than resolving it.

            \begin{quote} \textit{"I found ChatGPT hard to use as it was giving me solutions above my study level at the beginning."} Participant zelo-d6 \end{quote}
            
            Overall, the pre-experiment findings demonstrate that students already occupy a complex relationship with both formal academic support and generative AI systems. While they value independence and autonomy in their learning practices, they also experience anxiety and hesitation when seeking help through traditional channels. Simultaneously, students are already integrating AI tools into their learning workflows as a form of accessible, low-friction support. These findings therefore reinforce the rationale for the development of a course-specific Retrieval-Augmented Generation system capable of providing contextualised, module-aligned assistance while reducing barriers associated with conventional help-seeking behaviours

        \subsubsection{Post-Experiment Questionnaire}
            
            Post-experiment questionnaire responses demonstrated generally positive perceptions of Beacon across usability, relevance, learning support, and accessibility-related measures. Students frequently described Beacon as intuitive (73.3\%) and easy to use (73.35\%), with many participants agreeing that they felt confident navigating the interface and interacting with Beacon (66.7\%). Respondents indicated that they would use Beacon again for module support (33\% neutral; 46.6\% agree) and would recommend it to other students (33\% neutral; 40\% agree). However, the findings also identified recurring usability concerns, particularly regarding response speed, information density, and interface layout. Multiple students described Beacon’s outputs as overly long or visually overwhelming, suggesting that future versions should offer clearer formatting, collapsible sections, summarised responses, or visual distinctions between explanation categories. Despite these criticisms, the overall tone of responses suggested that students viewed Beacon as both promising and practically useful within a learning context with the majority having a positive experience using Beacon (33\% neutral; 60\% agree).
    
            Key to positive perceptions was Beacon’s use of module-specific materials. Many participants identified this as Beacon’s most valuable characteristic, emphasising that responses felt grounded in the actual content, terminology, and expectations of their modules. When compared with general-purpose AI systems such as \textit{ChatGPT}, several students noted that unrestricted systems often generated solutions that exceeded their level of study or provided complete code solutions without supporting understanding. In contrast, Beacon was praised for producing pseudocode, scaffolded explanations, and guided problem-solving that encouraged students to engage more critically with problems rather than simply copying solutions. Numerous students highlighted that Beacon “forced” them to think through solutions independently while still providing useful clarification and direction. When asked about the impact of Beacon on their learning, 53.35\% of students agreed Beacon improved their understanding of programming concepts, 53.3\% felt more confident solving programming problems and 60\% felt more confident tackling difficult topics. Consequently 66.7\% indicated that Beacon supported rather than replaced learning, while 80\% agreed that Beacon helped them learn independently. These findings therefore support the intended pedagogical design of Beacon. Rather than being a solution generator Beacon promotes independent learning and the development of learner confidence and autonomy. 

            \begin{quote} \textit{"traditional generative AI (such as ChatGPT or Gemini) rush to give you a full answer. Resultably, I dont learn a thing from it, just the result. Beacon however forces me to consider facts, but work out the answer myself."} Participant feno-j5 \end{quote}
            
            Perceptions of trust and relevance were generally positive, although participants demonstrated cautious rather than unconditional confidence in Beacon. Many agreed that Beacon’s answers were relevant (73.3\%), and aligned with module materials (93.3\%), while the inclusion of referenced source material significantly increased trust in the generated outputs (86.7\%). Students frequently reported that grounding responses in module-specific content made Beacon feel more academically reliable than unrestricted GenAI tools. Nevertheless, trust remained conditional, with most participants indicating that they would still verify important information with lecturers (40\% neutral; 46.7\% agree) or official course resources. Several students also reported occasional inaccuracies, hallucinations, or confusing responses, particularly when Beacon lacked sufficient contextual information within the stored teaching materials.

            \begin{quote} \textit{"It is very rigid in the sense of the second you of out of scope or ask something somewhat unrelated it becomes slightly useless."} Participant raku-n3 \end{quote}

            When asked about Beacons influence on help-seeking behaviours many students agreed that Beacon made it easier to seek help privately (86.7\%), reduced anxiety when unsure about a topic (53.3\%), and could be particularly beneficial for students with low confidence (93.3\%) or concerns about judgement (80\%). Several participants specifically noted that Beacon reduced barriers associated with embarrassment or reluctance to ask questions publicly. Students frequently highlighted the value of immediate support outside normal staff availability, positioning Beacon as a low-friction support mechanism that students could access without fear of negative evaluation. Importantly, while many participants still viewed tutors as the most effective source of help for complex or nuanced issues, Beacon was often described as a valuable “first step” before approaching teaching staff. This reinforces the broader argument that AI-supported systems may complement rather than replace traditional academic support structures.

            \begin{quote} \textit{"Beacon was helpful because it provided immediate support and made it easier to ask questions that I might otherwise feel hesitant or embarrassed to ask in class."} Participant noku-h7 \end{quote}
            
            Despite broadly positive feedback, positive experiences of Beacon were not universal. Some participants identified several important limitations and risks associated with Beacon. One of the most common criticisms concerned response speed, with many students comparing Beacon unfavourably to commercial AI systems that produce near-instant outputs. In some cases, students described Beacon struggling with very small code snippets, generating excessively detailed explanations for relatively simple problems, or producing different solutions to the same question depending on prompt phrasing. Additionally, a minority felt responses were overcomplicated, or failed to improve their understanding compared to asking a tutor directly. 

            \begin{quote} \textit{"small problems should have small answers. Especially when the target audience is people who are too shy to ask a teacher or somewhere in that ballpark. Having a small issue and getting a massive response can be overwhelming and make them feel further behind than they are."} Participant taro-x5 \end{quote}
            
            Although many students believed that Beacon promoted learning more effectively than unrestricted generative AI tools, some participants worried that students could still become dependent on Beacon or bypass independent critical thinking altogether. These responses indicate that the effectiveness of Beacon may vary according to individual learning preferences, confidence levels, and prior experience. Additionally, these concerns align with broader debates surrounding the educational risks of AI-supported learning technologies and reinforce the need for careful pedagogical integration and transparent system design.

        \subsubsection{Student Interviews}

            Analysis of the student interviews identified four key themes that compliment the findings from the pre- and post-study questionnaires. 

            Firstly, course-grounded guidance enhances the tools learning relevance. Students valued Beacon as it restricted responses to the scope of the module. This course-specifc grounding could prevent overly complex or irrelevant explanations that can be produced by a general-purpose AI. In contrast, Beacon was perceived being able to better explain concepts at an appropriate level. 

            \begin{quote} \textit{"With Beacon, I really found it useful that it was actually based on the exact knowledge we were taught in class."} Student Interview 3 \end{quote}
        
            In turn this reduced the need for students to navigate explanations that assumed greater prior knowledge or introduced unfamiliar concepts. Its ability to direct students to relevant slides, sessions, and examples also reduced the time and knowledge required to locate appropriate academic resources via a virtual learning environment. Furthermore, the 24/7 availability of Beacon was acknowledged as a benefit for students who lacked peer-support networks or needed assistance outside scheduled teaching hours. 

            Secondly, the psychological safety afforded by Beacon reduces help-seeking hesitation. In the interviews several students described feeling embarrassed or nervous about asking questions and acknowledged being anxious about revealing that they may be struggling with problems perceived as basic or fundamental. 

            \begin{quote} \textit{"I think for me, it was very useful because as you know, I never coded before. And some of the questions, like what I was thinking of asking in the first year, I felt it was embarrassing. Like, oh my God, I don't even know this, you know."} Student Interview 3 \end{quote}

            Beacon was viewed as being able to provide a private and non-judgemental first point of support, which could make help-seeking more accessible to students who lacked confidence or were reluctant to approach teaching staff. However, rather than merely replace tutors, students suggested Beacon acted as a bridge, allowing them first to develop confidence before seeking tutor support. This therefore suggests that Beacon can support help-seeking behaviour rather than eliminate it.
        
            \begin{quote} \textit{"it's kind of encouraging more questions, I think, which would eventually, I think, would enhance knowledge and marks because people would be more free to ask something like Beacon rather than go to you."} Student Interview 3 \end{quote}

            Thirdly, trust emerged through course-grounding and transparency. Students perceived Beacon's responses as trustworthy because they could see the source of information. This trust was further reinforced by the course-grounding of the materials. As Beacon stayed within the module boundaries students trusted responses reflected what was taught in-class and therefore felt confident that the information provided was consistent with module expectations and could support them in meeting assessment requirements.

            \begin{quote} \textit{"Beacon would know roughly more of the scope and how we've worked. And I think that would be really good for like creating the ideas [...] Because when you give a brief to an AI, it doesn't. It's not always the best. I would hope that Beacon, because it has more of the resources, would understand, oh, okay, we're going to do this and it's going to use this."} Student Interview 5 \end{quote}

            Finally, across the interviews a pedagogical tension between scaffolded learning and information convenience emerged. Students found value in Beacon's attempt to encourage learning, rather than provide direct answers.
        
            \begin{quote} \textit{"So something like ChatGPT would just go, here's the answer. Whereas Beacon basically makes you do a bit more work and gives you some of the steps towards reaching that answer rather than just being like, here you go."} Student Interview 4 \end{quote}

            However, students also cited this as a point of frustration, especially when Beacon would be overly restrictive or refuse to give answers beyond the scope of the module. Students suggested that this creates a risk where students who are time-pressured, or not motivated to engage with learning the material, would circumvent Beacon and use less restrictive tools.
    
            \begin{quote} \textit{"Students who will use the quick route, which happen regularly anyway, would typically just use one of the mainstream language models that can just give you the answer, simply because there's little to no friction to using it in comparison to Beacon where you actually have to work things out."} Student Interview 4 \end{quote}
    
            This friction is compounded by user-interface and user-experience issues students identified when using Beacon. The responses given by Beacon are highly templated which results in lengthy responses - both in terms of time and word count - even if the question is short. Furthermore, the templated nature of the responses restricts Beacon's ability to have conversational style interactions which are the norm in mainstream AI tools. These issues risked responses having unnecessary complexity which may in turn limit Beacons ability to support those in most need of assistance.
    
            \begin{quote} \textit{"it would come back with a with like the most full-on response it possibly can with like all of these different steps to the answer. But it's like, just give me a quick answer. If I want more information, let me ask for more information. Don't give it all to me at once."} Student Interview 1 \end{quote}
    
            These issues represent an ongoing challenge in the design of similar tools. A clear tension exists between providing sufficient information to demonstrate the value and encourage engagement in the learning process, versus withholding too much information so students circumvent the intervention in favour of mainstream AI tools that provide the answer rather than support learning. 
    
            In summary, across the five interviews Beacon's was viewed positively for providing private, timely, appropriately scoped, and pedagogically constrained assistance. However, its capacity to address disparities depended on careful calibration. Excessive guidance could enable task completion without learning, while excessive restriction, lengthy responses, and inflexible presentation could introduce new barriers. The findings therefore suggest that equitable course-specific RAG requires a balance between accessibility, productive cognitive effort, conversational adaptability, and continued access to human and peer support.

            Overall, the findings of the post-experiment questionnaires and interviews suggest that Beacon successfully occupied a middle ground between independent learning and formal academic support. Students valued Beacon’s ability to provide context-specific, module-aligned assistance while still encouraging active engagement with problems. The findings also indicate that students perceived Beacon as more pedagogically supportive than unrestricted generative AI tools because it constrained outputs within the boundaries of module expectations and promoted scaffolded learning rather than answer substitution. However, the results demonstrate that trust remained conditional, students continued to value human tutors for deeper support, and significant usability refinements are necessary before broader deployment. Taken together, these findings provide evidence that course-specific Retrieval-Augmented Generation systems may help reduce barriers to academic support while reinforcing independent study practices within HE computing contexts.
    
    \subsection{Academics' Views}

        Academic staff, like the students, saw value in Beacon as a scaffolded learning companion. In particular, the use of pseudocode, simple starting structures and links to relevant module materials were highlighted for providing structured learning guidance, rather than direct answers. This approach was considered appropriate for novice programmers because it encouraged students to translate guidance into working syntax and learn through implementation.

        \begin{quote} \textit{I think the main benefit to Beacon is definitely the explanation, the way that it can clearly explain step by step what everything means compared to other sort of chat bots or AI tools that will simply kind of give you the answer [...] and sometimes the AI expects you to know certain knowledge, but this is where sort of beacon is. It has the step up to it, it can actually explain core fundamentals of certain concepts."} Staff Interview 1 \end{quote}

        Scaffolding was further aided by Beacons grounding in module materials, which academic staff viewed as being a key mechanism in building trust in the responses. Consequently, staff felt Beacon could provide a more focused and trustworthy alternative to online searches and general-purpose AI, which may expose novice students to irrelevant, overly advanced, or pedagogically inappropriate information. Furthermore, this curriculum grounding was considered valuable in ensuring Beacon reflected what tutors intended students to learn and was able to communicate the practices and perspectives valued within the course rather than reproducing the average or dominant position found across internet sources.  
        
        \begin{quote} \textit{it's focused on code lab resources, so that you know, it's not going to lead you astray into a bunch of really confusing stuff, that's great, right? [...] Like there's just the way in which when you're trying to learn something, it's really helpful to not be trying to deal with a synthesis of everything that's ever been said by anyone anywhere on the internet."} Staff Interview 2 \end{quote}

        Regarding help-seeking, academic staff also viewed Beacon as an accessible gateway to academic support. Beacon was deemed beneficial for students who may have missed class, need concepts repeated, lacked confidence or require support outside normal teaching hours. 

        \begin{quote} \textit{"I think this is probably most useful for students who you know, didn't get it the first time or weren't there the first time maybe, and so kind of need a bit of repeat explanation to sort of work through it."} Staff Interview 2 \end{quote}

        However, rather than replace the tutor, Beacon was viewed as being complementary to tutor support and could allow students to seek initial guidance independently before escalating unresolved questions to teaching staff. This compliments the students views of Beacon providing a safe private space to develop confidence before seeking tutor support. 

        \begin{quote} \textit{"I think this is one really good step for students to be able to go, okay, let's try and learn it through this tool, see how far I can get with it with the clear explanations. And then if they don't get it, they can go one step higher."} Staff Interview 1 \end{quote}

        Although staff saw the positive benefits of Beacon, they also remained cautious and had concerns it may allow students to bypass the learning process. Staff noted unease with the ability of existing AI tools to give direct answers and the need for Beacon to support knowledge development rather than replace it by generating solutions. 

        \begin{quote} \textit{"this is a good one because I think ChatGPT does a lot for you. I think people are skipping that development stage."} Staff Interview 3 \end{quote}

        To this end staff valued Beacon because it resisted giving answers, but similar to student sentiments
        felt the nature of the responses were sometimes too rigid. The templated nature of responses meant staff noted the answers sometimes felt unnatural, lacked concision, or were unable to provide a natural conversational style experience typical of other available tools. Additionally, an ability to manipulate responses was noted, meaning it was on occasion possible to elicit a more direct answer.

        \begin{quote} \textit{"when kind of probing it to give some actual code, I think at first it gave the pseudo code and then I asked it to adapt it in a certain way and then it did give the C++ code"} Staff Interview 1 \end{quote}

        Staff also noted that whilst it knows the context of the module, it does not know the students context or level of knowledge (e.g. struggling with the fundamentals, or already beyond module scope). Staff emphasised that access to accurate information does not necessarily constitute accessible or educationally appropriate support. Lengthy and technically dense responses could increase the cognitive demands placed on students who were already confused. Conversely, inconsistent guardrails could allow students to obtain complete or assessment-relevant code without undertaking the intended learning. Responses therefore needed to be sensitive to the student's current knowledge, the type of question being asked and the relevant stage of the curriculum. 

        \begin{quote} \textit{"this is the thing that I think people struggle with. How do you give explanations that make sense to people who don't already get it? And I think I think the LLM suffer a little bit from [...] You know, they always want to give you the Wikipedia style. Here is this thing explained by people who know what it is."} Staff Interview 1 \end{quote}

        Thus, Beacon in its current form was noted as being best suited for those in the middle range regarding skill and understanding

        \begin{quote} \textit{"I think it would be generally useful for the students kind of in the middle, maybe a little bit on the earlier side, to just kind of get some more familiarity and just kind of be more immersed in some of the language as well."} Staff Interview 4 \end{quote}

        To make Beacon more adaptable staff suggested adding options for personalisation, either via on-screen settings or onboarding  questions so Beacon could gain an understanding of the individual learner with responses then tailored accordingly.

        \begin{quote} \textit{"if someone can just choose on a toggle, like, you know, explain like I’m five mode or I get it mode, that's interesting [...] you know, pitch the response. Maybe there's something even to say that as time goes on, like literally as the module is progressing, the prompt changes, right?."} Staff Interview 2 \end{quote}

        Finally, staff also highlighted an inherent tension between educational intent and student expectations (or desires). While restricting Beacon to taught content can preserve learning outcomes and prevent students from using advanced techniques prematurely, this same restriction could cause frustration among students who want a quick answer and thus abandon the system in favour of unrestricted generative AI.

        \begin{quote} \textit{I think putting on several different student hats, I can definitely see this as a good way to learn sort of step by step in getting the understanding of a concept for sure. I think putting on another student hat, I can see how it's frustrating that it just doesn't give the answer straight away."} Staff Interview 1 \end{quote}

        Overall, staff interviews, like the students, positioned Beacon as an intermediate layer in the help-seeking process, providing timely access to trusted course resources before students escalated difficulties to teaching staff. Academics identified particular value for students with limited prior knowledge, those who had missed teaching, and those reluctant to ask questions in class. Its curriculum grounding could reduce the navigational and evaluative burden associated with online resources while protecting novice learners from overly advanced, irrelevant or poor-quality information. However, overall educational value depends on adaptive scaffolding, persistent guardrails, cognitively accessible presentation, transparent curriculum boundaries and clear routes to human assistance. Beacon should therefore supplement rather than replace academic and peer support.

    \subsection{Summary of Findings}

        Taken together, the findings provide answers to the studies four guiding research questions.
        
        \subsubsection{RQ1: How do students perceive the usefulness and relevance of a course‐specific RAG system for academic support?}

        Students particularly valued Beacon's ability to explain concepts at the level of the module and direct them towards relevant slides, sessions, and examples. This contrasted with search engines, online tutorials, and general-purpose generative AI, which could assume greater prior knowledge, introduce concepts beyond the curriculum, or provide complete solutions without supporting understanding. However, while Beacon's attempt to preserve active learning was seen as useful, the findings revealed inconsistency in how effectively this was achieved. Some participants believed that Beacon introduced productive friction by requiring students to work towards a solution. Others found that it could be prompted to provide substantial or complete code, potentially enabling task completion without meaningful engagement. Persistent guardrails are therefore required across the full conversation, rather than only within the initial response.

        Usability was also a consistent concern. Lengthy and templated responses sometimes made accurate information difficult to locate or understand. This is especially significant because students seeking support may already be confused or have limited confidence in the subject. Participants recommended concise initial answers, clearer presentation of code and pseudocode, direct links to source materials, and optional expansion of explanations. More conversational follow-up responses and greater student control over response length, format, and level of technical detail could improve cognitive accessibility.

        \subsubsection{RQ2: To what extent do students perceive Beacon as reducing barriers associated with academic help‐seeking?}
        
        Beacon was perceived as a useful first step in the help-seeking process with findings suggesting that Beacon is best understood not simply as an AI tutor, but as an intermediary form of academic support between independent study and formal help-seeking. Across the questionnaires and interviews, participants valued the ability to access private, immediate, and module-aligned guidance while continuing to regard lecturers and tutors as important sources of support for more complex or nuanced difficulties. The value of Beacon therefore appeared to lie not in replacing human support, but in lowering the threshold for initiating help-seeking and enabling students to develop sufficient understanding either to continue independently or to seek more focused assistance from teaching staff. Beacon therefore extends the support ecosystem by providing a private and module-aligned first point of assistance. This is important because disparities in academic support are not only produced by whether support exists, but also by whether students feel able to access it.

        Although this study did not collect demographic data linking help-seeking behaviour to specific learner characteristics (Section~\ref{sec:limitations}), the mechanism identified here has plausible relevance to the demographic disparities that generative AI in education is often expected to address. Confidence and anxiety driven avoidance of help-seeking has been associated with gender \citep{huang2013gender, brown2021barriers}, first-generation and widening-participation status \citep{koh2022self, li2023college}, mature-student status \citep{chapman2017using}, and disability and neurodivergence \citep{anselimus2026generative}, among other factors. Generative AI tools more broadly carry both the potential to narrow and the risk of widening these disparities, depending on how they are designed and deployed \citep{ni2026mapping, james2024levelling, adejumo2026systematic}. Beacon's design - offering private, low-friction, course-aligned support outside the scrutiny of staff-mediated or public learning environments - targets precisely the anxiety- and confidence-related mechanisms most frequently implicated in these disparities. This suggests a plausible, though as yet untested, pathway by which course-specific RAG systems could narrow rather than widen educational disparities for the groups this special issue is concerned with; establishing whether this pathway holds in practice is the central task of the demographic-stratified evaluation proposed in Section~\ref{sec:future_work}.

        \subsubsection{RQ3: How do students evaluate the trustworthiness of responses generated by a system grounded in course materials?}

        The grounding of Beacon in lecturer-created materials was deemed central to enhancing the trustworthiness of responses. This trust emerged from transparency. As information sources were clearly attributed to lecture slides and leture material was often directly cited students felt the response were accurate and trusted explanations were consistent with the module's intended level and learning outcomes. 

        \subsubsection{RQ4: How do academic staff perceive Beacon’s value in reducing barriers to student help‐seeking, and what tensions does this raise for maintaining pedagogical safeguards and scaffolded learning}

        Staff shared student views regarding Beacon's ability to reduce help-seeking barriers. Benefits were perceived for students who lacked prior programming experience, had missed teaching, struggled to locate relevant resources, were working outside scheduled contact hours, or felt embarrassed about asking questions they perceived as basic. Scaffolded learning, curriculum grounding and resistance to providing direct answers were seen as key to this value, yet concerns were raised about the amount of friction these safeguards imposed. Students seeking rapid solutions might bypass Beacon in favour of unrestricted generative AI, while excessive restriction could prevent the system from providing sufficiently useful assistance. Therefore an ongoing challenge is maintaining the balance between pedagogical safeguards alongside student expectations from fast low-friction AI support. 
        
        Overall, the findings suggest that course-specific RAG has the potential to reduce help-seeking disparities associated with prior knowledge, confidence, attendance, resource navigation, and the availability of teaching staff. However, equitable access depends on more than making assistance continuously available. The support must also be cognitively accessible, responsive to individual learning needs, appropriately scaffolded, and integrated with existing academic and peer-support relationships. Beacon's potential therefore lies in balancing four requirements: lowering the threshold for initiating help-seeking, maintaining productive cognitive effort, adapting assistance to the learner and question, and providing clear routes to human support when automated guidance is insufficient.

        The findings therefore offer some practical insight for the design of course-specific RAG systems. Success depends less on retrieval accuracy and more on how effectively they balance curriculum grounding, trust, scaffolded learning and psychological support. Through curriculum-grounding responses become more useful and relevant to the students context and more accurately reflect the tutor intent and intended learning outcomes. Consequently, this leads to greater trust in the responses, especially when information sources are transparently attributed to provide a platform to find out more and review existing study materials. Furthermore, design should prioritise guidance over direct answers to scaffold learning, though careful balance is required to avoid too much friction and meet student expectations for convenience. Finally, by providing a private space to seek help, these systems offer psychological safety that can influence help-seeking behaviour. Thus, the design of these systems should augment not replace human-support by including psychological safeguards for adaptive guidance to meet different learner skills and encourage human escalation for persistent or complex problems. 
        
\section{Limitations}
    \label{sec:limitations}
    
    This study has several limitations. The evaluation was conducted within a single course context at a single institution, and involved a relatively small number of participants. Consequently, the findings may not generalise to other disciplines, institutions, or student populations.

    Beacon was also evaluated as a prototype rather than as part of a fully integrated learning management environment. Although participants were provided task scenarios designed to reflect real-world usage, it is acknowledged that these do not fully capture the range of questions students may ask. Furthermore, these tasks were conducted in a single sitting by each participant, thus usage does not reflect on-going interaction that would be expected if deployed in a live module setting. Consequently, the perceived usefulness and relevance of the tool may not accurately reflect its actual impact on module outcomes or its ability to address students' real-time learning needs.

    A related limitation concerns the verification of response accuracy. Evidence of Beacon's reliability in this study relied on self-reported student and staff perceptions rather than independent verification of Beacon's outputs. Although participants reported specific instances of hallucination and guardrail circumvention, these were identified anecdotally rather than systematically quantified. Future evaluations would benefit from combining perception-based measures with an independent assessment of response accuracy.

    The study also did not examine differences in system use or perceived benefit across specific learner groups, such as students with disabilities, neuro-divergent students, students from different socioeconomic backgrounds, or students with varying language backgrounds. As a result, while the findings suggest that Beacon may reduce some barriers to help-seeking, further research is required to determine whether such systems reduce educational disparities in practice and for whom they are most effective.

    Finally, this study did not evaluate Beacon's potential to reduce academic misconduct arising from unrestricted use of general-purpose generative AI, despite this being one of the original motivations for  development. Beacon was evaluated with students after their module had already been completed, rather than during live, assessed coursework, so there was no opportunity to observe whether its use influenced submitted work or the incidence of misconduct cases. This remains an anticipated rather than demonstrated benefit of the system.

\section{Future Work}
    \label{sec:future_work}

    Future work will explore the scalability of Beacon across multiple modules and investigate how such tools might be integrated into institutional learning platforms. While this study focused on a specific computing context, a key next step is to examine whether the same RAG approach can be applied across different modules, levels of study, and subject areas. This would require consideration of how module materials are collected, structured, updated, and indexed so that responses remain accurate and aligned with current teaching content. Scaling would also involve evaluating whether Beacon can maintain response quality when drawing from larger and more diverse knowledge bases, particularly where modules differ in terminology, assessment design, and expected student prior knowledge. Integration into institutional learning platforms would allow students to access support alongside lecture slides, workshop tasks, assessment briefs, and other approved resources. This could reduce barriers to use by embedding AI-supported assistance within the existing learning ecosystem rather than requiring students to access a separate tool.

    Additional research is also required to examine the long-term impact of AI-supported academic assistance on student learning outcomes. The present study captured students' immediate perceptions of Beacon following use, but it does not establish whether sustained engagement leads to measurable improvements in confidence, conceptual understanding, assessment performance, or help-seeking behaviour over time. Future longitudinal studies could investigate how students use the system across an entire module or academic year, whether usage patterns change as students become more confident, and whether Beacon encourages greater engagement with module materials. Such research would also help determine whether AI-supported assistance complements existing academic support structures or whether there is a risk of students becoming overly reliant on automated guidance. Examining long-term outcomes would therefore provide a clearer understanding of the pedagogical value of course-specific RAG systems and their role within HE learning support.

    Future work should also examine whether Beacon supports different student groups equitably, including students with lower confidence, students with disabilities, neuro-divergent students, students from widening participation backgrounds, and students with different levels of prior programming experience. This would allow future research to move beyond general perceptions of usefulness and examine whether course-specific AI support can measurably reduce disparities in access to academic help.

    Finally, future work should examine whether course-specific RAG systems such as Beacon can measurably reduce academic misconduct arising from unrestricted generative AI use, given the potential pathway from out-of-scope AI-generated content to misconduct proceedings discussed in Section \ref{sec:sys_design}. This would require deployment during a live, assessed module, comparing misconduct referral rates, or the proportion of flagged submissions attributable to out-of-scope AI-generated content, before and after the system's introduction.

\section{Conclusion}
    \label{sec:conclusion}

    The principal contribution of this study extends beyond the technical evaluation of a course-specific RAG system. Rather than viewing generative AI solely as a productivity tool or a potential source of academic misconduct, this study demonstrates how AI can be designed to address educational disparities by improving access to timely, contextually relevant academic support. Students who may otherwise hesitate to ask questions due to anxiety, fear of judgement, lack of confidence, or limited staff availability can instead access immediate guidance that remains aligned with module expectations. In this way, Beacon provides the potential for a more equitable model of learning while preserving opportunities for independent thinking and meaningful engagement with course materials.
    
    As GenAI becomes increasingly embedded within HE, the question is no longer whether students will use AI to support their learning, but how institutions can shape that use responsibly and equitably. This study argues that course-specific Retrieval-Augmented Generation offers one promising approach. By grounding AI responses within approved educational resources and positioning AI as a complement rather than a replacement for educators, institutions may be able to reduce barriers to help-seeking while maintaining trust, academic integrity, and pedagogical quality. Ultimately, the greatest opportunity for educational AI may lie not in replacing existing support structures, but in making them more accessible to the students who need them most.

\bmsection*{Author contributions}

    \textbf{Andy Gray:} 
    Writing – review \& editing, Writing – original draft, Conceptualisation, Software, Project administration, Methodology, Investigation, Formal analysis, Data curation, Funding acquisition. 
    
    \textbf{Jake Hobbs:} 
    Writing – review \& editing, Writing – original draft, Data curation, Funding acquisition, Methodology, Investigation, Formal analysis, Project administration. 

\bmsection*{Acknowledgements}
    \label{sec:ack}

    For the purpose of Open Access, the author has applied a CC-BY public copyright licence to any Author Accepted Manuscript (AAM) version arising from this submission. All underlying data to support the conclusions are provided within this paper.

\bmsection*{Funding Information}
    \label{sec:funding}
    This project was conducted as a result of acquiring research funding from 
    Bath Spa Universities 
    Creativity \& Curiosity Projects Fund.

\bmsection*{Conflict of interest statement}
    The authors declare no potential conflict of interest.

\bmsection*{Ethics Statement}
    Ethics approval was obtained from the School of Design ethics committee at 
    Bath Spa University 
    (Research Ethics Approval Number: 
    \textbf{240226JH}).
    
\bmsection*{Consent to participate and publication}
    All participants completed a consent from before data collection took place. Participants consented to anonymised quotes being published.
    
\bmsection*{Data availability statement}   
    All the data presented in the manuscript were obtained with the participants' consent to be published. Full data relating to this study are available on request from the author(s).

\bmsection*{AI Use Declaration Statement}

    During the preparation of this work author(s) used Claude (Anthropic; Sonnet 5) to assist with drafting, summarising, proofreading, refining academic content and to help identify potentially relevant literature. In addition Microsoft Co-Pilot was used to support data analysis as described in the methodology section. After using these tools the author(s) reviewed and edited content as needed and assume full responsibility for the content of the manuscript.

\bibliography{wileyNJD-AMA}

\appendix


\section{Exploratory Pilot Study Questionnaire}
    \label{app:pilot_qs}

\subsection*{Understanding the System}
\begin{enumerate}
    \item How clearly did you understand how the RAG system works based on the demonstration?*
        \begin{itemize}
            \item Very clearly
            \item Somewhat clearly
            \item Neutral
            \item Somewhat unclearly
            \item Not at all
        \end{itemize}

    
    \item Did the system’s responses make sense in relation to the questions asked?*
        \begin{itemize}
            \item Always
            \item Most of the time
            \item Sometimes
            \item Rarely
            \item Never
        \end{itemize}

\end{enumerate}

\subsection*{Perceived Usefulness}
\begin{enumerate}
    \item How relevant were the system’s responses to the course materials?*
    \begin{itemize}
        \item Very relevant
        \item Somewhat relevant
        \item Neutral
        \item Somewhat irrelevant
        \item Very irrelevant
    \end{itemize}
    
    \item Do you think this system could help clarify assessment requirements?*
    \begin{itemize}
        \item Yes, definitely
        \item Yes, somewhat
        \item Neutral
        \item Not really
        \item No, not at all
    \end{itemize}
    
    \item Would you trust the system’s responses when seeking academic support?*
    \begin{itemize}
        \item Yes, completely
        \item Yes, somewhat
        \item Neutral
        \item Not really
        \item No, not at all
    \end{itemize}
\end{enumerate}

\subsection*{Impact on Help-Seeking \\ \& Learning}
\begin{enumerate}
    \item If this system were available to you, how likely would you be to use it before asking a lecturer?*
    \begin{itemize}
        \item Very likely
        \item Somewhat likely
        \item Neutral
        \item Somewhat unlikely
        \item Very unlikely
    \end{itemize}
    
    \item Do you think the system could reduce hesitation in seeking help for assessments?*
    \begin{itemize}
        \item Yes, significantly
        \item Yes, somewhat
        \item Neutral
        \item No impact
        \item It would make me less likely to seek help
    \end{itemize}
\end{enumerate}

\section{Main Study: Student Pre-Questionnaire}
    \label{app:student_pre-questionnaire} 
    
    Prior to using Beacon, participants completed the following questionnaire to establish their educational background, existing use of learning resources and generative AI tools, and attitudes towards help-seeking and learning programming. 
    
    \subsection*{Participant and Educational Background} 
        
        \begin{enumerate} 
            \item ID Number 
            \item Year of study 
            \item Course 
            \item How would you rate your current programming ability? 
            \item What grade did you achieve in CodeLab I? 
        \end{enumerate} 
    
    \subsection*{Existing Learning Resources and AI Use} 
    
        \begin{enumerate} 
            \item Which online tools do you use to aid your understanding of the programming concepts introduced in CodeLab? 
            \item Which module resources do you use to aid your understanding of CodeLab module content? 
            \item How do you use AI tools within CodeLab? 
            \item What types of questions or prompts do you ask when using AI tools? 
            \item Which resources do you find most helpful in aiding your understanding of programming concepts? 
            \item Please provide a reason for your response to the previous question. 
        \end{enumerate} 
    
    \subsection*{Help-Seeking and Learning Confidence} 
        
        Please indicate your agreement with the following statements using a five-point Likert response format: 
        
        \begin{quote} 
            1 = Strongly Disagree, \quad 2 = Disagree, \quad 3 = Neutral, \quad 4 = Agree, \quad 5 = Strongly Agree 
        \end{quote} 
        
        \begin{enumerate} 
            \item I feel comfortable asking lecturers for help. 
            \item I often hesitate to ask questions in class. 
            \item I worry about being judged when asking for help. 
            \item I prefer to work through problems on my own before asking for help. 
            \item I sometimes avoid asking for help even when I need it. 
            \item I feel confident when studying difficult topics. 
            \item I feel anxious when I do not understand a topic. 
            \item I feel comfortable admitting when I do not understand something. 
        \end{enumerate} 
    
    \subsection*{Frequency of AI Use and Help-Seeking Behaviour} 
    
        Participants were asked to indicate how often they engaged in the following behaviours: 
        
        \begin{enumerate} 
            \item Use generative AI tools (e.g., ChatGPT or similar) when working on programming tasks for this module. 
            \item Avoid asking a tutor for help, even when you feel unsure about something. 
            \item Use generative AI tools instead of asking a tutor for help.
        \end{enumerate}

\section{Main Study: Task Scenarios}
    \label{app:student_tasks}

    \subsection*{Task 1}
        Given the code below you may feel you should already understand what happens here. This is the kind of question students often avoid asking tutors because it feels basic. There is no expectation that you already know the answer.
    
        \begin{verbatim}
        int count;
        cout << count;
        \end{verbatim}
        Use Beacon to explore what this code does and why it behaves this way.
        \newline
    
        \noindent In your own words, write an explanation below of:
        \begin{itemize} 
            \item What output you would expect 
            \item Why that output may occur                   
        \end{itemize}

    \subsection*{Task 2}
        You encounter the following error message when compiling your code.
        
        \begin{verbatim}
        error: expected ';' before 'return'
        \end{verbatim}
    
        \noindent Compiler messages are a common source of stress and frustration. Use Beacon to help you understand:

        \begin{itemize} 
            \item What this error message is trying to tell you 
            \item What kind of mistake might have caused it                
        \end{itemize}

    \subsection*{Task 3}
        Your tutor asks you to explain a while loop to another student. You are still learning this yourself, and you are given the following vague explanation to improve:
        \newline
    
        \noindent \textit{“A while loop runs while something is true.”}
        \newline
    
        \noindent Use Beacon to help you develop a clearer, more helpful explanation of a while loop.

    \subsection*{Task 4}
        Your tutor gave feedback on your code in class. They mentioned that your askUser() function was poorly designed due to a bad use of recursion and that a loop should be used instead. You did not fully understand what they meant or how to change the code.
        
        You did not feel comfortable asking them to explain again.

        \begin{verbatim}
        #include <iostream> 
        using namespace std;
        
        void askUser() {
          int choice;
          cout << "Continue? (1 = Yes, 2 = No): ";
          cin >> choice;
        
          if (choice == 1) {
            cout << "Continuing...\n";
          } else if (choice == 2) {
            cout << "Exiting...\n";
          } else {
            cout << "Invalid input. Try again.\n";
            askUser();
          }
        }
        
        int main() {
          askUser();
        }
        \end{verbatim}

        \noindent Use Beacon to help you:
        \begin{itemize} 
            \item Understand what your tutor likely meant by their feedback
            \item Explore why using a loop might be more appropriate than recursion in this case             
        \end{itemize}
	
        \noindent Provide:
        \begin{itemize} 
            \item A rewritten version of askUser() or Pseudocode 
            \item A short written explanation of why this approach is clearer or safer             
        \end{itemize}

     \subsection*{Task 5}
        You’ve been given the following task to complete by your tutor.

        \vspace{1\baselineskip}
        
            \noindent\textbf{\textit{Playing with Strings}}
            
            \noindent \textit{Write a program that asks for a user's first name and last name separately. The program should pass these strings to a function which returns the users full name as a single string.}\newline
            
            \noindent \textit{Next create another function that replaces every a, e, i , o, u with the letter z and returns the converted string}\newline
            
            \noindent \textit{Create a final function that reverses the user's name and returns the reversed string.}

        \vspace{1\baselineskip}
        
        \noindent This task is intentionally larger than previous ones and may feel overwhelming at first. You may use Beacon to support your thinking and development and provide below:

        \begin{itemize} 
            \item Your current solution or partial solution 
            \item A short summary of how you used Beacon to develop your solution           
        \end{itemize}

    \subsection*{Task 6}
    Use Beacon and ask it the kinds of questions you would normally ask another AI system (for example, ChatGPT or similar tools) when working on programming tasks.
    \newline
    
    \noindent Ask as many or as few questions as you like.

\section{Main Study: Student Post-Questionnaire}
    \label{app:student_post-questionnaire}
    
    Participants completed the following questionnaire after using the Beacon AI support system. 
        
        \subsection*{Response Scale} 
            Unless otherwise stated, the following items were measured using a five-point Likert scale: 
            \begin{quote} 1 = Strongly Disagree, 
                \quad 2 = Disagree, 
                \quad 3 = Neutral, 
                \quad 4 = Agree, 
                \quad 5 = Strongly Agree 
            \end{quote} 
            
        \subsection*{Usability and User Experience} 
            Please indicate your agreement with the following statements: \begin{enumerate} 
                \item Beacon was easy to use. 
                \item The interface felt intuitive and straightforward. 
                \item I felt confident navigating and interacting with Beacon. \item Beacon responded in a timely manner. 
                \item I had a positive experience using Beacon. 
                \item I would like to use Beacon again for module support. \item I would recommend Beacon to other students.
            \end{enumerate} 
            
            Open-ended questions: 
            \begin{enumerate} 
                \setcounter{enumi}{7} 
                \item What features would you like to see added or improved in future versions of Beacon? 
                \item Please provide any additional comments that would help improve Beacon. 
            \end{enumerate} 
                
            \subsection*{Perceived Answer Quality and Trust} 
                    
                Please indicate your agreement with the following statements: 
                    
                    \begin{enumerate} 
                        \item Beacon's answers were clear and easy to understand. 
                        \item Beacon's answers were accurate and relevant to my query. 
                        \item Beacon made good use of module-specific materials. 
                        \item I trusted the information provided by Beacon. \item I felt confident using the answers in my learning. 
                        \item I understood where Beacon's answers came from. 
                        \item Seeing source material increased my trust in the responses. 
                        \item I would still verify important information with lecturers or course materials. 
                        \item I was concerned that the system might provide incorrect information. 
                    \end{enumerate} 
                    
                \subsection*{Errors and Hallucinations} 
                    
                    \begin{enumerate} 
                        \item How often did you receive incorrect or unhelpful responses? 
                        \begin{itemize} 
                            \item Never 
                            \item Rarely 
                            \item Sometimes 
                            \item Often 
                            \item Very Often 
                        \end{itemize} 
                        
                        \item If you encountered incorrect or confusing responses, please describe an example. 
                    \end{enumerate} 
                    
                \subsection*{Impact on Learning} 
                    
                    Please indicate your agreement with the following statements: 
                        \begin{enumerate} 
                            \item Using Beacon improved my understanding of programming concepts. 
                            \item I feel more confident in solving programming problems after using Beacon. 
                            \item Beacon helped me learn independently without needing staff support. 
                            \item Using Beacon increased my confidence in tackling difficult topics. 
                            \item Beacon supported my learning rather than replacing it. 
                        \end{enumerate} 
                    
                    Open-ended question: 
                    \begin{enumerate} 
                        \setcounter{enumi}{5} 
                        \item In what ways did Beacon support (or not support) your learning? 
                    \end{enumerate} 
                    
                \subsection*{Accessibility and Psychological Safety} 
                    
                    Please indicate your agreement with the following statements: 
                    \begin{enumerate} 
                        \item I felt more comfortable asking Beacon than asking a tutor. 
                        \item Beacon made it easier to seek help privately. 
                        \item Beacon reduced my anxiety when I was unsure about a topic. 
                        \item Beacon made academic support feel more accessible. 
                        \item Beacon provides useful support outside of normal staff availability. 
                        \item Beacon could be especially helpful for students with low confidence. 
                        \item Beacon could help students who feel judged when asking questions. 
                        \item Some students may rely on the system without thinking independently. 
                    \end{enumerate} 
                    
                \subsection*{Comparison with Alternative Support Sources}
                    Compared to other sources of support, please indicate your evaluation of Beacon: 
                    \begin{enumerate} 
                        \item Compared to online resources (e.g., Stack Overflow, YouTube), Beacon was: 
                        \item Compared to generative AI tools (e.g., ChatGPT, Microsoft Copilot), Beacon was: 
                        \item Compared to asking a tutor, Beacon was:
                    \end{enumerate} 
                    
                    Response scale: 
                    \begin{quote} Much Worse 
                        \quad | \quad Worse \quad | \quad About the Same \quad | \quad Better \quad | \quad Much Better \end{quote} Open-ended question: \begin{enumerate} \setcounter{enumi}{3} \item Please provide your reasoning for the responses given in the previous questions. \end{enumerate}

\section{Post Pilot Student Semi-Structured Interview Questions}
    \label{app:student_interviews}
    
    \begin{enumerate}
        \item What were your expectations of Beacon before you started using it?
        \item How would you describe your overall experience interacting with Beacon?
        \item Were there any moments where using Beacon felt particularly smooth or helpful?
        \item Did you encounter any difficulties or frustrations while using Beacon?
        \item How would you assess the accuracy and usefulness of Beacon's responses?
        \item Did using Beacon change how you approached your programming work or problem-solving?
        \item In what ways, if any, did Beacon support your learning during the pilot?
        \item How did Beacon compare to other support sources you typically use?
        \item Is there anything you would change about Beacon?
        \item Would you use Beacon again in future modules? Why or why not?
    \end{enumerate}

\section{Post Pilot Academic Semi-Structured Interview Questions}
    \label{app:staff_interviews}
    
    \begin{enumerate}
        \item What were your expectations of Beacon before you started using it?
        \item How would you describe your overall experience interacting with Beacon?
        \item Were there any moments where using Beacon felt particularly smooth or helpful?
        \item Did you encounter any difficulties or frustrations while using Beacon?
        \item How would you assess the accuracy and usefulness of Beacon's responses?
        \item Do you see Beacon being useful for students? How / Why?
        \item How did Beacon compare to other support sources you typically use?
        \item Is there anything you would change about Beacon?
        \item Would you use Beacon again in future modules? Why or why not?
    \end{enumerate}

\end{document}